\documentstyle[a4wide,11pt,epsf]{article}

\newcommand{\qit}[1]{{\sl ``#1''}}           
\newcommand{\sys}[1]{{\verb+#1+}}            
\newcommand{\qsys}[1]{{``\verb+#1+''}}       
\newcommand{\usr}[1]{{\sl #1}}               
\newcommand{\pipe}{\; | \;}
\newcommand{\mem}[1]{\mbox{{\em #1}}}        

\title{Natural Language Interfaces to Databases -- An
  Introduction\thanks{
  To appear in the {\em Journal of Natural Language
    Engineering}, Cambridge University Press. Also
  available as Research Paper no.\ 709, Department of Artificial
  Intelligence, University of Edinburgh, 1994.}}
\author{I. Androutsopoulos$^\dagger$ \and G.D.\ Ritchie$^\dagger$
        \and P.\ Thanisch$^\ddagger$} 
\date{$^\dagger$Department of Artificial Intelligence, 
      University of Edinburgh \\
      80 South Bridge, Edinburgh EH1 1HN, Scotland, U.K. \\
      e-mail: \verb+ion@aisb.ed.ac.uk+, \ \verb+G.D.Ritchie@ed.ac.uk+ 
      \vspace{3mm}\\
      $^\ddagger$Department of Computer Science, University of Edinburgh \\
      King's Buildings, Mayfield Road, Edinburgh EH9 3JZ, Scotland, U.K. \\
      e-mail: \verb+pt@dcs.ed.ac.uk+
     } 

\begin{document}
\maketitle

\begin{abstract}

  This paper is an introduction to natural language interfaces to
  databases ({\sc Nlidb}s). A brief overview of the history of {\sc
    Nlidb}s is first given. Some advantages and disadvantages of {\sc
    Nlidb}s are then discussed, comparing {\sc Nlidb}s to formal query
  languages, form-based interfaces, and graphical interfaces. An
  introduction to some of the linguistic problems {\sc Nlidb}s have to
  confront follows, for the benefit of readers less familiar with
  computational linguistics. The discussion then moves on to {\sc
    Nlidb} architectures, portability issues, restricted natural
  language input systems (including menu-based {\sc Nlidb}s), and {\sc
    Nlidb}s with reasoning capabilities. Some less explored areas of
  {\sc Nlidb} research are then presented, namely database updates,
  meta-knowledge questions, temporal questions, and multi-modal {\sc
    Nlidb}s. The paper ends with reflections on the current state of
  the art.

\end{abstract}

\section{Introduction}

A natural language interface to a database ({\sc Nlidb}) is a system
that allows the user to access information stored in a database by
typing requests expressed in some natural language (e.g.\ English).
The following example is a dialogue between the user and {\sc Loqui},
a commercially available {\sc Nlidb} \cite{Binot} (the dialogue is
borrowed from \cite{Loqui}; the system's responses are slightly
simplified). Throughout this paper messages printed by a {\sc Nlidb}
are shown in \sys{this typeface}; user entries are shown in \usr{this
  typeface}.

\smallskip

\noindent
\sys{>} \usr{Who works on 3 projects?} \\
\sys{B.\ Vandecapelle, C.\ Willems, D.\ Sedlock, J.L.\ Binot, L.\
  Debille, ...} \\ 
\sys{>} \usr{Which of them are project leaders?} \\
\sys{D.\ Sedlock, J.L.\ Binot}\\
\sys{>} \usr{Documents describing their projects?} \\
\sys{{\sc Bim}\_{\sc Loqui}: "The {\sc Loqui} {\sc Nlidb}", 
                             "{\sc Bim}\_{\sc Loqui}"} \\
\sys{{\sc Mmi}2: "Technical Annex"} \\
\sys{>} \usr{How many of these projects do not finish before 1994?} \\
\sys{2} \\
\sys{{\sc Bim}\_{\sc Loqui}, {\sc Mmi2}}\\
\sys{>} \usr{Are they led by JLB or DS?} \\
\sys{The former.} 
\smallskip

The purpose of this paper is to serve as an introduction to {\em
  some}\/ key concepts, problems, methodologies, and lines of research
in the area of natural language interfaces to databases. This paper is
by no means a complete discussion of all the issues that are relevant
to {\sc Nlidb}s. The reader is also referred to \cite{Copestake} and
\cite{Perrault} for alternative surveys of the same
field.\footnote{Sections \ref{nl_problems} and \ref{architectures}
  were greatly influenced by \cite{Perrault}.} A detailed description
of a natural language front-end that reflects the current state of the
art can be found in \cite{Alshawi} and \cite{Alshawi2}.

Although the paper contains hints about the capabilities of existing
{\sc Nlidb}s, it does not contain complete descriptions of particular
systems, nor is the purpose of this paper to compare particular {\sc
  Nlidb}s.  This paper is mainly based on information obtained from
published documents. The authors do not have personal hands-on
experience with most of the {\sc Nlidb}s that will be mentioned.
Whenever a system's feature is mentioned, this means that the
documents cited state that the particular system provides this
feature, and it is not implied that other systems do not have similar
capabilities.
Finally, this paper assumes that the user's requests are communicated
to the {\sc Nlidb} by typing on a computer keyboard. Issues related to
speech processing are not discussed.

The remainder of this paper is organised as follows: section
\ref{history} is a brief overview of the history of {\sc
  Nlidb}s\footnote{Section \ref{history} is largely based on
  information from \cite{Copestake}, \cite{Perrault}, and
  \cite{Templeton}.}; section \ref{comparison} discusses the
advantages and disadvantages of {\sc Nlidb}s; section
\ref{nl_problems} presents some of the linguistic problems {\sc
  Nlidb}s have to cope with; section \ref{architectures} describes
architectures adopted in existing {\sc Nlidb}s; section
\ref{portability} discusses portability issues related to {\sc
  Nlidb}s; section \ref{restricted} introduces {\sc Nlidb}s that
explicitly restrict the set of natural language expressions the user
is allowed to input, so that the user can have a clearer view of what
sorts of questions the system can understand; section
\ref{intelligent} describes {\sc Nlidb}s with reasoning modules;
section \ref{further_topics} highlights some less explored areas of
{\sc Nlidb} research, namely database updates, meta-knowledge
questions, temporal questions, and multi-modal interfaces; finally,
section \ref{state_of_the_art} summarises the current state of the
art.

\section{Some history} \label{history}

Prototype {\sc Nlidb}s had already appeared in the late sixties and
early seventies. The best-known {\sc Nlidb} of that period is {\sc
  Lunar} \cite{Woods}, a natural language interface to a database
containing chemical analyses of moon rocks.  {\sc Lunar} and other
early natural language interfaces were each built having a particular
database in mind, and thus could not be easily modified to be used
with different databases. (Although the internal representation methods
used in {\sc Lunar} were argued to facilitate independence between the
database and other modules \cite{Woods1968}, the way that these were
used was somewhat specific to that project's needs. Portability issues
are discussed in section \ref{portability}.)

By the late seventies several more {\sc Nlidb}s had appeared. {\sc
  Rendezvous} \cite{Codd1} engaged the user in dialogues to help
him/her formulate his/her queries. {\sc Ladder} \cite{Hendrix2} could
be used with large databases, and it could be configured to interface
to different underlying database management systems ({\sc Dbms}s).
{\sc Ladder} used semantic grammars (discussed in section
\ref{semantic_grammars}), a technique that interleaves syntactic and
semantic processing. Although semantic grammars helped to implement
systems with impressive characteristics, the resulting systems proved
difficult to port to different application domains.  Indeed, a
different grammar had to be developed whenever {\sc Ladder} was configured
for a new application.  As researchers started to focus on portable
{\sc Nlidb}s, semantic grammars were gradually abandoned. 
{\sc Planes} \cite{Waltz} and {\sc Philiqa1} \cite{Scha} were some of the other
{\sc Nlidb}s that appeared in the late seventies.

{\sc Chat-80} \cite{Warren} is one of the best-known {\sc Nlidb}s
of the early eighties. {\sc Chat-80} was implemented entirely in
Prolog. It transformed English questions into Prolog expressions,
which were evaluated against the Prolog database. The code of {\sc
  Chat-80} was circulated widely, and formed the basis of several other
experimental {\sc Nlidb}s (e.g.\ {\sc Masque} \cite{Auxerre1}
\cite{Auxerre2} \cite{Androutsopoulos3}).

In the mid-eighties {\sc Nlidb}s were a very popular area of research,
and numerous prototype systems were being implemented. A large part of
the research of that time was devoted to portability issues.  For
example, {\sc Team} \cite{Grosz1} \cite{Grosz2} \cite{Martin} was
designed to be easily configurable by database administrators with no
knowledge of {\sc Nlidb}s.

{\sc Ask} \cite{Thompson2} \cite{Thompson1} allowed end-users to teach the
system new words and concepts at any point during the interaction.
{\sc Ask} was actually a complete information management system, providing
its own built-in database, and the ability to interact with multiple
external databases, electronic mail programs, and other computer
applications. All the applications connected to {\sc Ask} were accessible to
the end-user through natural language requests. The user stated
his/her requests in English, and {\sc Ask} transparently generated suitable
requests to the appropriate underlying systems.

{\sc Janus} \cite{Hinrichs1} \cite{Resnik} \cite{Weischedel}
\cite{Bobrow2} had similar abilities to interface to multiple
underlying systems (databases, expert systems, graphics devices, etc).
All the underlying systems could participate in the evaluation of a
natural language request, without the user ever becoming aware of the
heterogeneity of the overall system. {\sc Janus} is also one of the
few systems to support temporal questions (discussed in section
\ref{temporal_questions}).

Systems that also appeared in the mid-eighties were {\sc
  Datalog}\footnote{This {\sc Nlidb} has nothing to do with the subset
  of Prolog described in \cite{Ullman}, which is also called
  {\sc Datalog}.} \cite{Hafner2} \cite{Hafner1}, {\sc Eufid}
\cite{Templeton}, {\sc Ldc} \cite{Ballard1} \cite{Ballard2}, {\sc Tqa}
\cite{Damerau2} \cite{Damerau}, {\sc Teli} \cite{Ballard3}, and many
others.

Although some of the numerous {\sc Nlidb}s developed in the
mid-eighties demonstrated impressive characteristics in certain
application areas, {\sc Nlidb}s did not gain the expected rapid and
wide commercial acceptance.  For example, in 1985 Ovum Ltd.
\cite{Johnson} (p.14) was foreseeing that \qit{By 1987 a natural
  language interface should be a standard option for users of {\sc
    Dbms} and `Information Centre' type software, and there will be a
  reasonable choice of alternatives.}.\footnote{A more recent report
  on natural language markets by the same company is \cite{Ovum1991}.}
Since then, several commercially available {\sc Nlidb}s have appeared,
and some of them are claimed to be commercially successful. However,
{\sc Nlidb}s are still treated as research or exotic systems, rather
than a standard option for interfacing to databases, and their use is
certainly not wide-spread. The development of successful alternatives
to {\sc Nlidb}s, like graphical and form-based interfaces, and the
intrinsic problems of {\sc Nlidb}s (both discussed in the following
section) are probably the main reasons for the lack of acceptance of
{\sc Nlidb}s.

In recent years there has been a significant decrease in the number of
papers on {\sc Nlidb}s published per year. Still, {\sc Nlidb}s
continue to evolve, adopting advances in the general natural language
processing field (e.g.\ discourse theories -- section
\ref{dialogue_systems}), exploring architectures that transform {\sc
  Nlidb}s into reasoning agents (section \ref{intelligent}), and
integrating language and graphics to exploit the advantages of both
modalities (section \ref{multimodal}), to name some of the lines of
current research.  Generic linguistic front-ends have also appeared.
These are general-purpose systems that map natural language input to
expressions of a logical language (e.g.\ the {\sc Cle} system
\cite{Alshawi} -- see also \cite{Alshawi2}). These generic front-ends
can be turned into {\sc Nlidb}s, by attaching additional modules that
evaluate the logic expressions against a database (section
\ref{architectures}).
The following are {\em some}\/ of the commercially available {\sc Nlidb}s:
\begin{itemize}
\item {\sc Intellect} \cite{Harris4} from Trinzic (formed by the merger of
  AICorp and Aion). This system is based on experience from {\sc Robot}
  \cite{Harris3} \cite{Harris2} \cite{Harris1}.

\item {\sc Bbn}'s {\sc Parlance} \cite{BBN}, built on experience from
  the development of the {\sc Rus} \cite{Bobrow1} and {\sc Irus}
  \cite{Bates2} systems.

\item {\sc Ibm}'s {\sc Languageaccess} \cite{Ott}. This system stopped being
  commercially available in October 1992.

\item Q\&A from Symantec.

\item {\sc Natural Language} from Natural Language Inc. According to
  \cite{Copestake}, this system was previously known as {\sc
    Datatalker}, it is described in \cite{Manferdelli}, and it is
  derived from the system described in \cite{Ginsparg}.

\item {\sc Loqui} \cite{Binot} from {\sc Bim}.

\item {\sc English Wizard} from Linguistic Technology
  Corporation. The company was founded by the author of AICorp's
  original {\sc Intellect}. 

\end{itemize}
\noindent 
Some aspects of the linguistic capabilities of {\sc Intellect}, Q\&A, and
{\sc Natural Language} are reviewed in \cite{Sijtsma}.

It should be noted that when some researchers use the term
``database'', they often just mean ``a lot of data''. In this paper,
we mean quite a lot more than that. Most importantly, the data is
assumed to represent some coherent attempt to model a part of the
world. Furthermore, the database is structured according to some model
of data, and the {\sc Nlidb} is designed to work with that data model.
Database systems have also evolved a lot during the last decades. The
term ``database system'' now denotes (at least in computer science)
much more complex and principled systems than it used to denote in the
past. Many of the underlying ``database systems'' of early {\sc
  Nlidb}s would not deserve to be called database systems with today's
standards.

In the early days of database systems, there was no concept of naive
end-users accessing the data directly; this was done by an expert
programmer writing a special computer program. The reason for this was
the `navigational' nature of the data model used by these early
database systems. Not only did the user need to know about the
structure of the data in the application.  He/she also needed to know
many programming tricks to get at the data.
The development of the relational model of data in the 1970's (Codd
\cite{Codd1970}) had a major impact on database systems.  In the
relational model, the only storage structure is the table, and this
was something that even naive users could understand. Relatively
simple declarative query languages, such as {\sc Sql} (see section
\ref{comparison}), were developed for this class of user.

Currently, there are two major developments in database technology
that will have an impact on {\sc Nlidb}s. The first is the growing
importance of object-oriented database systems, and the second is the
trend in relational database technology towards more complex storage
structures to facilitate advanced data modelling.  We note that both
of these trends could make it harder to produce an {\sc Nlidb}. They both
reflect a tendency to concentrate on new complex database application
areas, such as network management and computer-aided design, where the
user is anything but naive, and the immediate access to the database
will often be carried out by a layer of application software.

\section{Advantages \& disadvantages} \label{comparison}

This section discusses some of the advantages and disadvantages of
{\sc Nlidb}s, comparing them to formal query languages, form-based
interfaces, and graphical interfaces.

Access to the information stored in a database has traditionally been
achieved using formal query languages, such as {\sc Sql} (see
\cite{Ullman} for an introduction to {\sc Sql}). Let us assume, for
example, that the following tables are stored in a relational
database:

\begin{center}
{\small
\begin{tabular}{lr}
\begin{tabular}{lll}
\multicolumn{3}{c}{employees\_table} \\
employee & department & phone \\
\hline
Thompson & sales & 2317 \\
Richardson & accounting & 2554 \\
\dots & \dots & \dots 
\end{tabular}
&
\begin{tabular}{lll}
\multicolumn{3}{c}{departments\_table} \\
department & manager & city \\
\hline
sales & Ferguson & London \\
accounting & Richardson & Bristol \\
\dots & \dots & \dots 
\end{tabular}
\end{tabular}
} 
\end{center}

The first table, employees\_table, shows the department and phone
number of each employee. The second table, departments\_table, shows
the manager of each department, and the city where each department is
located. A list showing each employee's manager can be created using
the {\sc Sql} query:
\begin{center}
{\small
\begin{tabular}{l}
\sys{SELECT employees\_table.employee, departments\_table.manager} \\
\sys{FROM employees\_table, departments\_table} \\
\sys{WHERE employees\_table.department = departments\_table.department}
\end{tabular}
}
\end{center}

\noindent The {\sc Sql} query above instructs the database system to
report all possible pairs consisting of an employee value and a
manager value, such that the employee value comes from a row
in employees\_table, the manager value comes from a row in
departments\_table, and the two rows have the same department
value.

\smallskip

In form-based interfaces (e.g.\ \cite{Ingres1}), pre-defined forms
consisting of fields are used. The user fills the information which is
already known in the corresponding fields, and the system completes
the remaining fields by querying the underlying database.  A user
wishing to learn the manager of an employee named \qit{Thompson}\/
could fill in the pre-defined form ``\sys{Employee Information Form}''
as shown below on the left. The system would respond by filling the
remaining fields as shown below on the right.

\begin{center}
{\small
\begin{tabular}{lr}
\begin{tabular}{ll} 
\multicolumn{2}{c}{\sys{Employee Information Form}} \\
\hline
\sys{Employee:} & \usr{Thompson} \\
\sys{Department:} & \\
\sys{Phone:} & \\
\sys{Manager:} & 
\end{tabular}
&
\begin{tabular}{ll} 
\multicolumn{2}{c}{\sys{Employee Information Form}} \\
\hline
\sys{Employee:} & \sys{Thompson} \\
\sys{Department:} & \sys{sales} \\
\sys{Phone:} & \sys{2317} \\
\sys{Manager:} & \sys{Ferguson}
\end{tabular} 
\end{tabular}
} 
\end{center}

\noindent If there are two or more answers (e.g. more than one employee
called \qit{Thompson}\/), the system responds by generating a stack of
filled forms, one form per possible answer.

In a more powerful method, called {\em query by example} (proposed by
Zloof, see \cite{Ullman} for a description), the user can combine an
arbitrary number of forms, where each form reflects the structure of a
database table. The manager of \qit{Thompson} could be found using the
following query-by-example forms:

\begin{center}
{\small
\begin{tabular}{lr}
\begin{tabular}{lll}
\multicolumn{3}{c}{\sys{employees\_table\_form}} \\
\sys{employee} & \sys{department} & \sys{phone} \\
\hline
\usr{Thompson} & \usr{X} & 
\end{tabular}
&
\begin{tabular}{lll}
\multicolumn{3}{c}{\sys{departments\_table\_form}} \\
\sys{department} & \sys{manager} & \sys{city} \\
\hline
\usr{X} & \usr{P.Y} & 
\end{tabular}
\end{tabular}
} 
\end{center}

\noindent The first form shows that the system should select from
employees\_table rows having \usr{Thompson} as employee value.
The two occurrences of variable \usr{X}\/ show that each row selected
from employees\_table should be joined with rows from
departments\_table that have the same department values. The
\usr{P.Y}\/ entry shows that the corresponding manager value is
to be printed.

\begin{figure}
\begin{center}
\mbox{\epsfysize=1in \epsffile{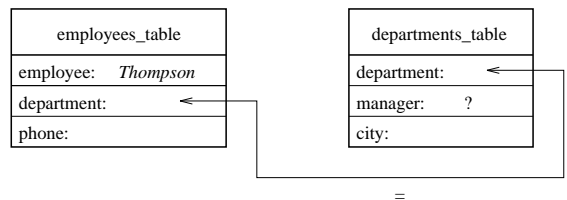}}
\caption{A graphical query}
\label{graph_query}
\end{center}
\end{figure}

Finally, in one approach to graphical interfaces (similar to the one
used in SunSimplify \cite{SunSimplify}) the user first selects the
database tables to be used in the query ($employee\_table$\/ and
$manager\_table$\/ in our example). Each selected table appears on
the screen as a frame consisting of attribute-slots. The user can then
fill in attribute slots by typing on the keyboard, he/she can join
attributes across the various frames using the mouse, or he/she can impose
restrictions on attributes using the mouse and menu options. To find
the manager of \qit{Thompson}, a user could use the graphical query
shown in figure \ref{graph_query}.
A similar method is to allow frames to correspond to world entities
(e.g.\ managers, employees, departments), rather than to database
tables.

\medskip

The following advantages and disadvantages of {\sc Nlidb}s are often
mentioned in the literature. 

\subsubsection*{Some advantages of NLIDBs}

\paragraph{No artificial language:} 

One advantage of {\sc Nlidb}s is supposed to be that the user is not
required to learn an artificial communication language.  Formal query
languages are difficult to learn and master, at least by
non-computer-specialists.  Graphical interfaces and form-based
interfaces are easier to use by occasional users; still, invoking
forms, linking frames, selecting restrictions from menus, etc.\ 
constitute artificial communication languages, that have to be learned
and mastered by the end-user. In contrast, an ideal {\sc Nlidb} would
allow queries to be formulated in the user's native language. This
means that an ideal {\sc Nlidb} would be more suitable for occasional
users, since there would be no need for the user to spend time
learning the system's communication language.

In practice, current {\sc Nlidb}s can only understand limited subsets
of natural language. Therefore, some training is still needed to teach
the end-user what kinds of questions the {\sc Nlidb} can or cannot
understand.  In some cases, it may be more difficult to understand
what sort of questions an {\sc Nlidb} can or cannot understand, than
to learn how to use a formal query language, a form-based interface,
or a graphical interface (see disadvantages below). One may also argue
that a subset of natural language is no longer a natural language.

\paragraph{Better for some questions:} 

It has been argued (e.g.\ \cite{Cohen}) that there are kinds of
questions (e.g.\ questions involving negation, or quantification) that
can be easily expressed in natural language, but that seem difficult
(or at least tedious) to express using graphical or form-based
interfaces.  For example, \qit{Which department has no programmers?}
(negation), or \qit{Which company supplies every department?}
(universal quantification), can be easily expressed in natural
language, but they would be difficult to express in most graphical or
form-based interfaces.  Questions like the above can, of course, be
expressed in database query languages like {\sc Sql}, but complex
database query language expressions may have to be written.

\paragraph{Discourse:} 

Another advantage of {\sc Nlidb}s, mentioned in \cite{Hendrix1} and
\cite{Cohen}, concerns natural language interfaces that support
anaphoric and elliptical expressions (sections \ref{anaphora} and
\ref{ellipsis}). {\sc Nlidb}s of this kind allow the use of very
brief, underspecified questions, where the meaning of each question is
complemented by the discourse context. In formal query languages,
graphical interfaces, and form-based interfaces this notion of
discourse context is usually not supported. This point will become
clearer in sections \ref{anaphora} and \ref{ellipsis}.

\subsubsection*{Some disadvantages of NLIDBs}

\paragraph{Linguistic coverage not obvious:} A frequent complaint against
{\sc Nlidb}s is that the system's linguistic capabilities are not
obvious to the user \cite{Tennant1} \cite{Hendrix1} \cite{Cohen}.  As
already mentioned, current {\sc Nlidb}s can only cope with limited
subsets of natural language. Users find it difficult to understand
(and remember) what kinds of questions the {\sc Nlidb} can or cannot
cope with.  For example, {\sc Masque} \cite{Auxerre2} is able to
understand \qit{What are the capitals of the countries bordering the
  Baltic {\em and} bordering Sweden?}, which leads the user to assume
that the system can handle all kinds of conjunctions (false positive
expectation).  However, the question \qit{What are the capitals of the
  countries bordering the Baltic {\em and} Sweden?}\/ cannot be
handled.  Similarly, a failure to answer a particular query can lead
the user to assume that ``equally difficult'' queries cannot be
answered, while in fact they can be answered (false negative
expectation).

Formal query languages, form-based interfaces, and graphical
interfaces typically do not suffer from these problems. In the case of
formal query languages, the syntax of the query language is usually
well-documented, and any syntactically correct query is guaranteed to
be given an answer. In the case of form-based and graphical
interfaces, the user can usually understand what sorts of questions
can be input, by browsing the options offered on the screen; and any
query that can be input is guaranteed to be given an answer.

\paragraph{Linguistic vs.\ conceptual failures:} When the {\sc Nlidb}
cannot understand a question, it is often not clear to the user
whether the rejected question is outside the system's {\em
  linguistic}\/ coverage, or whether it is outside the system's {\em
  conceptual}\/ coverage \cite{Tennant1}.  Thus, users often try to
rephrase questions referring to concepts the system does not know
(e.g.\ rephrasing questions about salaries towards a system that knows
nothing about salaries), because they think that the problem is caused
by the system's limited linguistic coverage. In other cases, users do
not try to rephrase questions the system could conceptually handle,
because they do not realise that the particular phrasing of the
question is outside the linguistic coverage, and that an alternative
phrasing of the same question could be answered.  Some {\sc Nlidb}s
attempt to solve this problem by providing diagnostic messages,
showing the reason a question cannot be handled (e.g.\ unknown word,
syntax too complex, unknown concept, etc.) (see section
\ref{response_generation}).

\paragraph{Users assume intelligence:} As pointed out in \cite{Hendrix1},
{\sc Nlidb} users are often misled by the system's ability to process
natural language, and they assume that the system is intelligent, that
it has common sense, or that it can deduce facts, while in fact most
{\sc Nlidb}s have no reasoning abilities (see however section
\ref{intelligent}). This problem does not arise in formal query
languages, form-based interfaces, and graphical interfaces, where the
capabilities of the system are more obvious to the user.

\paragraph{Inappropriate medium:} As mentioned in \cite{Binot}, it has been
argued that natural language is not an appropriate medium for
communicating with a computer system.  Natural language is claimed to
be too verbose or too ambiguous for human-computer interaction.  {\sc
  Nlidb} users have to type long questions, while in form-based
interfaces only fields have to be filled in, and in graphical
interfaces most of the work can be done by mouse-clicking.  Natural
language questions are also often ambiguous, while formal, form-based,
or graphical queries never have multiple meanings.  Section
\ref{nl_problems} presents some of the problems in this area, and
discusses how {\sc Nlidb}s attempt to address them.

\paragraph{Tedious configuration:} {\sc Nlidb}s usually require tedious and
lengthy configuration phases before they can be used (see section
\ref{portability}). In contrast, most commercial database systems have
built-in formal query language interpreters, and the implementation of
form-based interfaces is largely automated.

\medskip

Several experiments have been carried out, comparing how users cope
with formal query languages, form-based interfaces, graphical
interfaces, and {\sc Nlidb}s. 

\cite{Bell} describes one such experiment, during which
fifty five subjects, ranging from computer novices to programmers,
were asked to perform database queries using a formal query language
({\sc Sql}), a graphical interface (SunSimplify -- see section
\ref{comparison}), and a {\sc Nlidb} ({\sc Datatalker}, later known as
{\sc Natural Language} -- see section \ref{history}). The subjects
first received some training on how to use the three interfaces, and
were then asked to perform database queries, most of which were
similar to the queries the subjects had encountered during the
training period.  The experiment measured the number of queries the
subjects managed to perform successfully, and the average time the
subjects used to perform each successful query.
None of the three interfaces could be said to be a winner. Each
interface was better in some kinds of queries, and in most queries the
subjects' performance was roughly the same, whether the subjects were
asked to use {\sc Sql}, the graphical interface, or the {\sc Nlidb}. Roughly
speaking, the {\sc Nlidb} seemed to be better in queries where data from many
tables had to be combined, and in queries that were not similar to the
ones the users had encountered during the training period.
Information about similar experiments can be found in \cite{Jarke} and
\cite{Small}. 

\cite{Whittaker} discusses various approaches that have been used in
the evaluation of natural language systems, and describes an
experiment where first a Wizard of Oz was used to collect sample user
questions, and then the sample questions were used as input to an
actual {\sc Nlidb}. In a Wizard of Oz experiment, the user interacts
with a person who pretends to be an {\sc Nlidb} through a computer
network. The user is not aware that he/she is not interacting with a
real {\sc Nlidb}. \cite{Martin1981} provides a set of questions
towards a {\sc Nlidb} that was collected using a Wizard of Oz
experiment. \cite{Diaper1986} describes in detail a Wizard of Oz
experiment.

\cite{Capindale1990} provides information about several experiments on
the usability of {\sc Nlidb}s, and describes in detail one such
experiment that assessed the usability of {\sc Intellect} (see section
\ref{history}). The authors conclude that ``{\em natural language is
  an effective method of interaction for casual users with a good
  knowledge of the database, who perform question-answering tasks, in
  a restricted domain}''. \cite{Dekleva1994} also describes an
experiment on the usability of {\sc Intellect}, and concludes that
{\sc Nlidb}s are a practical alternative.

\section{Linguistic problems} \label{nl_problems}

This section presents {\em some}\/ of the linguistic problems a {\sc
  Nlidb} has to confront when attempting to interpret a natural
language request. Most of these problems do not arise only in {\sc
  Nlidb}s; they are common in most kinds of natural language
understanding systems.
It is stressed that this section is by no means a full list of the
linguistic problems {\sc Nlidb}s have to confront. This section merely
presents some of the problems, mainly for the benefit of readers not
familiar with computational linguistics. 

The reader is reminded that messages generated by the {\sc Nlidb} are
printed in \sys{this typeface}, and that user entries are printed in
\usr{this typeface}.

\subsection{Modifier attachment} \label{modif_att}

Let us consider the request {\em ``List all employees in the company with a 
driving licence''}. Linguistically, both {\em ``in the company''}\/ and 
{\em ``with a driving licence''}\/ are {\em modifiers}, i.e. they modify the 
meaning of other syntactic constituents. The problem is to identify the 
constituent to which each modifier has to be attached.

A human would immediately understand that {\em ``in the company''}\/ refers to
{\em ``employees''}, and that {\em ``with a driving licence''}\/ refers to the
{\em ``employees in the company''}.
This reading of the request corresponds to the structure:
\begin{center}
{\em List all [ [employees [in the company]] [with a driving licence] ].}
\end{center}
and can be paraphrased as {\em ``From the employees working in the
  company, list those having a driving licence''}.  However, a
computer system would also consider the following reading:
\begin{center}
{\em List all [employees [in the [company [with a driving licence]]]].}
\end{center}
where {\em ``with a driving licence''}\/ refers to the {\em
  ``company''}.  The latter reading can be paraphrased as {\em ``List
  all employees working in the company which has a driving licence''}.
This second reading cannot be easily ruled out. For example, in the
similar request {\em ``List all employees in the company with an
  export licence.''}, it is the second reading that probably has to be
chosen (\qit{with an export licence} refers to \qit{the company}).  To
be able to choose the correct reading, the system must know that in
the particular application domain employees can have driving
licences but not export licences, and that companies can have export
licences but not driving licences.

Alternatively, some systems attempt to resolve modifier attachment
ambiguities by using heuristics.  {\sc Pre} \cite{Epstein} uses a
``most right association principle'' to resolve relative clauses
attachment: relative clauses are assumed to modify the rightmost
available constituent. For example, in \qit{List an employee who was
  hired by a recruiter whose salary is greater than \$3,000.}, the
relative clause \qit{whose salary is greater than \$3,000}\/ is
assumed to refer to the \qit{recruiter}, not the \qit{employee}. More
advanced disambiguation techniques for prepositional phrase attachment
can be found in \cite{Whittemore}.

Still, in some cases modifier attachment is truly ambiguous. Consider
the request (borrowed from \cite{Perrault}) {\em ``List all employees
  in the division making shoes''}. In some application domains {\em
  ``making shoes''} may equally well refer to the {\em ``division''}\/
or to the {\em ``employees''}. In such cases, a {\sc Nlidb} can (a)
present to the user paraphrases of both readings, and ask the user to
chose one of the readings, or (b) it can print answers corresponding
to both readings, indicating which answers refer to which readings.

\subsection{Quantifier scoping}

Determiners like \qit{a}, \qit{each}, \qit{all}, \qit{some}\/ are
usually mapped to logic quantifiers. In cases of sentences with many
words of this kind, it is difficult to determine which quantifier
should be given a wider scope.
Let us consider the question \qit{Has every student taken some course?}. The
question has two readings. ($student$\/ and $course$\/ are variables.)
\begin{enumerate}

\item Check that\\ $\forall student \; 
                     \exists course \;
                        taken(student, course)$

\item Check that\\ $\exists course \;
                     \forall student \;
                        taken(student, course)$

\end{enumerate}
In the first reading, each student is allowed to have taken a different course, 
while in the second reading all students must have taken the same course.

A possible heuristic method to resolve scoping ambiguities
\cite{Perrault} is to prefer scopings preserving the left-to-right
order of the quantifiers.  In \qit{Has every student taken some
  course?}, \qit{every}\/ appears before \qit{some}. Therefore,
according to the heuristic, the first of the readings above should be
preferred, since in this reading the universal quantifier introduced
by \qit{every}, precedes the existential quantifier introduced by
\qit{some}.

Another approach \cite{Perrault} is to associate a numeric strength to
each determiner.  In this approach, determiners with greater strengths
are given wider scopes. Returning to the question \qit{Has every
  student taken some course?}, if the strength of \qit{every}\/ is
greater than the strength of \qit{some}, then the first reading is
again preferred.  The reader is referred to chapter 8 of
\cite{Alshawi} for a detailed discussion of methods that can be used
to resolve quantifier scoping problems.

\subsection{Conjunction and disjunction}

The word \qit{and} is often used to denote disjunction rather than
conjunction.  This introduces ambiguities which are difficult to
resolve. The following two examples, borrowed from \cite{Templeton},
illustrate the problem:

In \qit{List all applicants who live in California and Arizona.}\/
every applicant who lives {\em either} in California or Arizona should
be reported. The possibility for \qit{and} to denote a conjunction
should be ruled out, since an applicant cannot normally live in more
than one state.  {\sc Eufid} \cite{Templeton} detects cases where it
is conceptually impossible for \qit{and} to denote a conjunction, and
turns \qit{and}s to \qit{or}s. (Section 4.2 of \cite{Wallace1984}
describes a heuristic rule for detecting cases where \qit{and} cannot
have a conjunctive meaning.)

\qit{Which minority and female applicants know Fortran?}, however, could mean
either \qit{Which female applicants belonging to a minority know Fortran?}\/ or
\qit{Which applicants who are female or belong to a minority know Fortran?}.
In such cases \qit{and} is truly ambiguous. 

As in the case of {\sc Eufid}, when asked \qit{How many people live in
  Belmont and Boston?}, {\sc Parlance} \cite{BBN} understands that
\qit{and} cannot denote a conjunction, and reports the population of
each city.  In contrast, when asked \qit{What's the average salary of
  blacks and hispanics?}, {\sc Parlance} provides answers
corresponding to both possible meanings, by generating the following
tables:

\begin{center}
{\small
\begin{tabular}{lr}
\begin{tabular}[t]{lll}  
Average Salary & Ethnic Group & Count \\
\hline 
\$42,481.81 & Black & 11 \\
\$45,230.00 & Hispanic & 10 
\end{tabular}  
&
\begin{tabular}[t]{lll}  
Average Salary & Count \\
\hline
\$43,790.47 & 21 
\end{tabular} 
\end{tabular}
} 
\end{center}

\subsection{The nominal compound problem}

In English, nouns are often modified by other preceding nouns. The
meaning of the resulting compound constituent is hard to predict. The
following examples, based on examples from \cite{Perrault}, illustrate
the problem:

\begin{itemize}

\item \qit{city department} could denote a department located in a
  city, or a department responsible for a city.

\item \qit{research department} is probably a department {\em carrying
    out}\/ research.

\item \qit{research system} is probably a system {\em used in}\/
  research, it is not a system carrying out research.

\end{itemize}

Similar difficulties arise in the case of adjective-noun compounds.
For example, a \qit{large company}\/ may be a company with a large
volume of sales or a company with many employees, a \qit{big
  department} may be a department with many employees or a department
occupying a lot of space, etc.

Because of the difficulty in determining the meaning of noun-noun and
adjective-noun compounds from the meanings of the individual
compounded nouns or adjectives, some {\sc Nlidb}s require the meaning
of each possible noun-noun or adjective-noun compound to be declared
during the configuration phase (see section \ref{portability}).  In
{\sc Ldc-1} \cite{Ballard1} \cite{Ballard2}, for example, whenever the
system is taught a new noun (e.g.\ \qit{department}\/), the configurer
is asked for all adjectives (e.g.\ \qit{large}, \qit{small},
\qit{profitable}\/) and nouns (e.g.  \qit{sales}, \qit{city}\/) that
can precede the new noun as modifiers.  The configurer is then asked
to define the meaning of each possible compound (\qit{large
  department}, \qit{profitable department}, \qit{sales department},
etc.) in terms of concepts of the underlying database.

\subsection{Anaphora} \label{anaphora}

Pronouns (e.g.\ \qit{she}, \qit{they}\/), possessive determiners (e.g.\
\qit{his}, \qit{their}\/), and noun phrases (e.g.\ \qit{the project},
\qit{these people}\/) are often used to denote implicitly entities
mentioned in the discourse, a linguistic phenomenon called {\em anaphora}.
Consider the following dialogue between the user and {\sc Ask} from
\cite{Thompson2}:

\smallskip
\begin{tabular}{l}
\sys{>} \usr{Is there a ship whose destination is unknown?} \\
\sys{yes} \\
\sys{>} \usr{What is it?} \\
\sys{What is [the ship whose destination is unknown]?} \\
\sys{Saratoga}
\end{tabular}
\smallskip

\noindent {\sc Ask} has understood that \qit{it}\/ refers to \qit{the
  ship whose destination is unknown}.  {\sc Ask} echoes back the
user's question, with the pronoun replaced by what the system
understands to be its meaning.  This prevents the user from being
misled, in cases where the pronoun anaphora is not resolved correctly.

The following is a (shortened) dialogue with {\sc Loqui} from
\cite{Loqui}:

\smallskip
\begin{tabular}{l}
\sys{>} \usr{Who leads {\sc TPI}?} \\
\sys{E.\ Feron} \\
\sys{>} \usr{Who reports to him?} \\
\sys{C.\ Leonard, C.\ Willems, E.\ Bidonnet, P.\ Cayphas, J.P.\ Van Loo} \\ 
\sys{>} \usr{What do they work on?} \\
\sys{ \begin{tabular}{ll}
      project & \ worker \\
      \hline 
      {\sc Docdis} & \begin{tabular}[t]{l}
               C.\ Willems \\
               J.P.\ Van Loo \\
               P.\ Cayphas
               \end{tabular}
               \\
      {\sc Eurs} & \begin{tabular}[t]{l}
              C.\ Leonard \\
              C.\ Willems \\
              E.\ Bidonnet
                   \end{tabular}
      \end{tabular}}\\
\sys{>} \usr{Which of these are leaders?} \\
\sys{J.P.\ Van Loo}
\end{tabular}
\smallskip

A simplistic method to handle pronoun anaphora is to keep a list of all
entities mentioned in the discourse. When a pronoun is encountered, the
system examines the list, usually starting from the most recent
entries, and associates each pronoun with the most recently mentioned
entity that satisfies grammatic and semantic constraints.

In \qit{Who reports to him?}\/ above, a system could reason that
\qit{him}\/ refers to \qit{E.\ Feron}\/ rather than to the project
\qit{{\sc Tpi}}, because the pronoun is of masculine gender, while a
project is of neutral gender.  In \qit{Which of these are leaders?}\/
above, {\sc Loqui} has probably reasoned that \qit{these}\/ refers to
the people mentioned in the previous response, rather than to the
projects mentioned in the previous response, using information stating
that only persons not projects can be leaders.

Possessive determiner anaphora (e.g. \qit{their}\/) is similar to 
pronoun anaphora. The following example is also from \cite{Loqui}:

\smallskip
\begin{tabular}{l}
\sys{>} \usr{Who reports to MM?} \\
\sys{E.\ Feron, J.L.\ Binot} \\
\sys{>} \usr{What are their projects?} \\
\sys{ \begin{tabular}{ll}
      project & person \\
      \hline
      {\sc Bim}\_{\sc Loqui} & J.L.\ Binot \\
      {\sc Eurs} & E.\ Feron
      \end{tabular} } 
\end{tabular}
\smallskip

Noun phrases can also be used as anaphoric expressions. In the
following dialogue, from \cite{Loqui}, \qit{the project}\/ is an
anaphoric noun phrase referring to the project \qit{{\sc Mmi2}}:

\smallskip
\begin{tabular}{l} 
\sys{>} \usr{Who leads {\sc Mmi2}?} \\
\sys{J.L.\ Binot} \\
\sys{>} \usr{Who works on the project?} \\
\sys{B.\ Vandecapelle, D.\ Sedlock, F.\ Balfroid, J.L.\ Binot, L.\ Debille}
\end{tabular}
\smallskip

Anaphora is a well-known problem in natural language processing. See
\cite{Hirst} for an overview of the issues, and \cite{Hobbs1986} for
relatively simple methods that can be used to resolve many cases of
pronoun anaphora.

\subsection{Elliptical sentences} \label{ellipsis}

In every-day discourse people use incomplete (elliptical) sentences,
the meaning of which is complemented by the discourse context. This
phenomenon is illustrated in the following dialogue (from \cite{BBN})
between the user and {\sc Parlance} :

\smallskip
\noindent 
\sys{>} \usr{Does the highest paid female manager have any degrees
  from Harvard?}\\ 
\sys{Yes, 1.}\\
\sys{>} \usr{How about {\sc Mit}?}\\
\sys{No, none.} \\
\sys{>} \usr{Who is the manager of the largest department?} \\
\sys{ \begin{tabular}[t]{lll}
      Name & Dept. & Count \\
      Patterson & 045 & 40 
      \end{tabular} } \\
\sys{>} \usr{The smallest department?} \\
\sys{ \begin{tabular}[t]{lll}
      Name & Dept. & Count \\
      Saavedra & 011 & 2
      \end{tabular} } 
\smallskip

\noindent \qit{How about {\sc Mit}?} and \qit{The smallest department?} are
elliptical sentences.

Handling complex cases of elliptical sentences requires a discourse
model to be maintained by the system (see section
\ref{dialogue_systems}).  {{\sc Xcalibur} \cite{Carbonell} uses a set of
\qit{discourse expectation}\/ and \qit{contextual substitution}\/
rules, that allow elliptical utterances, such as the ones shown below, to be
understood:

\smallskip
\begin{tabular}{l}
\sys{>} \usr{What is the price of the three largest single port fixed
  media disks?} \\ 
\sys{>} \usr{speed?} \\
\sys{>} \usr{two smallest?} \\
\sys{>} \usr{How about the price of the two smallest?} \\
\sys{>} \usr{Also the smallest with two ports?} \\
\sys{>} \usr{speed with two ports?} 
\end{tabular}
\smallskip

The handling of elliptical sentences is a desirable feature in a {\sc
  Nlidb}.  {\sc Nlidb} users typically do not have single stand-alone
questions to ask.  The first, often exploratory query, is usually
followed by many elliptical follow-up queries. Some {\sc Nlidb}s (e.g.
{\sc Chat-80} \cite{Warren}) are oriented towards the processing of
stand-alone questions, providing no support for elliptical questions.
In such systems, the questions shown above would have to be formulated
as:

\smallskip
\begin{tabular}{l}
\sys{>} \usr{What is the price of the three largest single port fixed
  media disks?} \\ 
\sys{>} \usr{What is the speed of the three largest single port fixed
  media disks?} \\ 
\sys{>} \usr{What is the speed of the two smallest single port fixed
  media disks?} \\ 
\sys{>} \usr{What is the price of the two smallest single port fixed
  media disks?} \\ 
\sys{>} \usr{What is the price of the two smallest fixed media disks
  with two ports?} \\ 
\sys{>} \usr{What is the speed of the two smallest fixed media disks
  with two ports?}  
\end{tabular}
\smallskip

\noindent Having to type every question in full can be annoying for
the user, though the problem can be alleviated by providing a facility
for editing previously entered questions.

\subsection{Extragrammatical utterances}

Most linguistic theories describe the structure and meaning of
grammatical, perfectly formed utterances. However, every-day language
often contains misspellings, syntactically ill-formed input,
telegraphic utterances, etc. If the main goal of a {\sc Nlidb} is
to assist the user, then the system must be able to understand the user's
requests, even when they are partially ill-formed. 

Most of the existing {\sc Nlidb}s can handle only simple cases of
extragrammatical utterances. For example, {\sc Parlance} forgives dropped
articles, sentences where subject and verb cases or numbers do not
agree, capitalisation errors and punctuation errors \cite{BBN}. It is,
thus, able to understand sentences like \qit{whats reynolds
  departments name}, or \qit{people in dePT 45 who is supervisory
  scientist}.  {\sc Ask} uses its dictionary to correct spelling errors and
errors caused by missing spaces \cite{Thompson1}. In \qit{What is the
  crago of the Orient Clipper?} it converts \qit{crago}\/ to
\qit{cargo}, and in \qit{What is the cargoof the Orient Clipper?}\/ it
converts \qit{cargoof}\/ to \qit{cargo of}.

\section{Architectures} \label{architectures}

\subsection{Pattern-matching systems} \label{pattern_matching}

Some of the early {\sc Nlidb}s relied on pattern-matching techniques
to answer the user's questions. To illustrate a simplistic
pattern-matching approach, consider a database table holding
information about countries:

\begin{center}
\begin{tabular}{lll}
\multicolumn{3}{c}{{\sc Countries\_table}} \\
{\sc Country} & {\sc Capital} & {\sc Language} \\
\hline
France & Paris & French \\
Italy & Rome & Italian \\
\dots & \dots & \dots 
\end{tabular}
\end{center}
A primitive pattern-matching system could use rules like:

\begin{center}
\begin{tabular}{ll}
pattern: & \dots \qit{capital} \dots $<$country$>$ \\
action : & Report {\sc Capital} of row where {\sc Country} =
$<$country$>$\\
&\\
pattern: & \dots \qit{capital} \dots \qit{country} \\
action : & Report {\sc Capital} and {\sc Country} of each row
\end{tabular}
\end{center}

\noindent The first rule says that if a user's request contains the word
\qit{capital}\/ followed by a country name 
(i.e. a name appearing in the {\sc Country} column), then the system 
should locate the row which contains the country name, 
and print the corresponding capital.

If, for example, the user typed \qit{What is the capital of Italy?},
the system would use the first pattern-rule, and report \qsys{Rome}.
The same rule would allow the system to handle \qit{Print the capital
  of Italy.}, \qit{Could you please tell me what is the capital of
  Italy?}, etc. In all cases the same response would have been
generated.

According to the second rule, any user request containing the word
\qit{capital}\/ followed by the word \qit{country} should be handled
by printing the capital of each country, as listed in the database
table.  For example, \qit{What is the capital of each country?},
\qit{List the capital of every country.}, \qit{Capital and country
  please.}, etc., would all be handled by the second rule.

The main advantage of the pattern-matching approach is its simplicity: no
elaborate parsing and interpretation modules (see later sections) are
needed, and the systems are easy to implement. Also,
pattern-matching systems often manage to come up with some reasonable answer, 
even if the input is out of the range of sentences the patterns were designed 
to handle. Returning to the example above, the second rule would allow the
system to answer the question \qit{Is it true that the capital of each country 
is Athens?}, by listing the capital of each country, which can be considered
as an indirect negative answer.

Pattern-matching systems are not necessarily based on such simplistic
techniques as the ones discussed above. {\sc Savvy}, a pattern
matching system discussed in \cite{Johnson} (p.153), employs
pattern-matching techniques similar to the ones used in signal
processing.

According to \cite{Johnson}, some pattern-matching systems were able
to perform impressively well in certain applications. However, the
shallowness of the pattern-matching approach would often lead to bad
failures.  In one case (mentioned in \cite{Johnson}), when a
pattern-matching {\sc Nlidb} was asked \qit{{\sc titles of employees
    in los angeles.}}, the system reported the state where each
employee worked, because it took \qit{{\sc in}}\/ to denote the post
code of Indiana, and assumed that the question was about employees and
states.

\subsection{Syntax-based systems} \label{syntax_based}

In syntax-based systems the user's question is parsed (i.e. analysed
syntactically), and the resulting parse tree is directly mapped to an
expression in some database query language.  A typical example of this
approach is {\sc Lunar} \cite{Woods}.

Syntax-based systems use a grammar that describes the possible
syntactic structures of the user's questions. The following example,
based on a similar example from \cite{Perrault}, shows an
over-simplistic grammar in a {\sc Lunar}-like system.

\[
\begin{array}{lll}
S & \rightarrow & \mem{NP} \; \; \;  \mem{VP} \\
\mem{NP} & \rightarrow & \mem{Det} \; N  \\
\mem{Det} & \rightarrow & \mbox{\qit{what}} \; \pipe \; \mbox{\qit{which}} \\
N & \rightarrow & \mbox{\qit{rock}} \; \pipe \; \mbox{\qit{specimen}}
   \; \pipe \; \mbox{\qit{magnesium}} \; \pipe \mbox{\qit{radiation}}
   \; \pipe \; \mbox{\qit{light}} \\ 
\mem{VP} & \rightarrow & V \; \; \;  N \\
V & \rightarrow & \mbox{\qit{contains}} \; \pipe \; \mbox{\qit{emits}}
\end{array}
\]

\begin{figure}
\begin{center}
\mbox{\epsfysize = 1.1in \epsffile{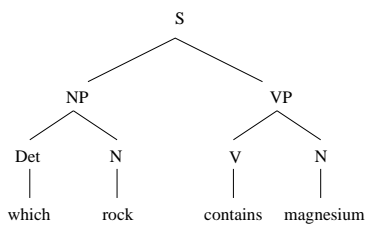}}
\caption{Parse tree in a syntax-based system}
\label{lunar_tree}
\end{center}
\end{figure}

The grammar above says that a sentence ($S$\/) consists of a noun
phrase ($\mem{NP}$\/) followed by a verb phrase ($\mem{VP}$\/), that a
noun phrase consists of a determiner ($\mem{Det}$\/) followed by a
noun ($N$\/), that a determiner may be \qit{what}\/ or \qit{which},
etc.  Using this grammar, a {\sc Nlidb} could figure out that the
syntactic structure of the question \qit{which rock contains
  magnesium}\/ is as shown in the parse tree of figure
\ref{lunar_tree}.
The {\sc Nlidb} could then map the parse tree of figure \ref{lunar_tree} to
the following database query (\sys{X} is a variable):

\begin{verbatim}
(for_every X (is_rock X)
             (contains X magnesium) ;
             (printout X))
\end{verbatim}

\noindent which would then be evaluated by the underlying database
system. This mapping would be carried out by rules, and would be
completely based on the syntactic information of the parse tree. The
system of our example could use mapping rules stating that:

\begin{itemize}

\item The mapping of \qit{which}\/ is \sys{for\_every X}.

\item The mapping of \qit{rock}\/ is \sys{(is\_rock X)}.

\item The mapping of an {\em NP}\/ is $Det' \; N'$, where $Det'$ and
  $N'$ are the mappings of the determiner and the noun respectively.
  Thus, the mapping of the {\em NP}\/ subtree in our example is
  \sys{for\_every X (is\_rock X)}.

\item The mapping of \qit{contains}\/ is \sys{contains}.

\item The mapping of \qit{magnesium}\/ is \sys{magnesium}.

\item The mapping of a {\em VP}\/ is \sys{(}$V'$ \sys{X} $N'$\sys{)},
  where $V'$ is the mapping of the verb, and $N'$ is the mapping of
  the noun-sister of the verb. Thus, the mapping of the {\em VP}\/
  subtree in our example is \sys{(contains X magnesium)}.

\item The mapping of an S is \sys{(}$\mem{NP}\,' \; \mem{VP}\,'$
  \sys{; (printout X))}, where $\mem{NP}\,'$, $\mem{VP}\,'$ are the
  mappings of the {\em NP}\/ and {\em VP}\/ subtree accordingly.
  Thus, in our example, the mapping of the sentence is as shown above.

\end{itemize}

Syntax-based {\sc Nlidb}s usually interface to application-specific
database systems, that provide database query languages carefully
designed to facilitate the mapping from the parse tree to the database
query.  It is usually difficult to devise mapping rules that will
transform directly the parse tree into some expression in a real-life
database query language (e.g.\ {\sc Sql}).

\subsection{Semantic grammar systems} \label{semantic_grammars}

In semantic grammar systems, the question-answering is still done by 
parsing the input and mapping the parse tree to a database query.
The difference, in this case, is that the grammar's categories
(i.e.\ the non-leaf nodes that will appear in the parse tree) do not
necessarily correspond to syntactic concepts. Figure
\ref{sem_gram_fig} shows a possible semantic grammar.

\begin{figure}
{\small
\[
\begin{array}{l}
S \;\;  \rightarrow \;\; Specimen\_question \pipe \;\; Spacecraft\_question \\
Specimen\_question \;\; \rightarrow \;\; Specimen \;\;
Emits\_info \pipe Specimen \;\; Contains\_info \\
Specimen \;\; \rightarrow \;\; \mbox{\qit{which rock}} \pipe
\mbox{\qit{which specimen}} \\ 
Emits\_info \;\; \rightarrow \;\; \mbox{\qit{emits}} \;\;  Radiation \\
Radiation \;\; \rightarrow \;\; \mbox{\qit{radiation}} \pipe
\mbox{\qit{light}} \\ 
Contains\_info \;\; \rightarrow \;\; \mbox{\qit{contains}} \;\; Substance \\
Substance \;\; \rightarrow \;\; \mbox{\qit{magnesium}} \pipe
\mbox{\qit{calcium}} \\ 
Spacecraft\_question \;\; \rightarrow \;\; Spacecraft \;\;
Depart\_info \pipe Spacecraft \;\; Arrive\_info \\
Spacecraft \;\; \rightarrow \;\; \mbox{\qit{which vessel}} \pipe
\mbox{\qit{which spacecraft}} \\ 
Depart\_info \;\; \rightarrow \;\; \mbox{\qit{was launched on}}
\;\; Date \pipe \mbox{\qit{departed on}} \;\; Date \\ 
Arrive\_info \;\; \rightarrow \;\; \mbox{\qit{returns on}} \;\; Date
\pipe \mbox{\qit{arrives on}} \;\; Date  
\end{array}
\]
\caption{A semantic grammar}
\label{sem_gram_fig}
}
\end{figure}

The reader will have noticed that some of the categories of the
grammar of figure \ref{sem_gram_fig} (e.g.\ $Substance$, $Radiation$,
$Specimen\_question$\/) do not correspond to syntactic constituents
(e.g.\ noun-phrase, noun, sentence).  Semantic information about the
knowledge domain (e.g.\ that a question may either refer to specimens
or spacecrafts) is hard-wired into the semantic grammar.
Semantic grammar categories are usually chosen to enforce semantic 
constraints. For example, the grammar above does not allow \qit{light}\/
to follow the verb \qit{contains}. (In contrast, the grammar of the
previous section allows \qit{contains light}.)

The grammar's categories can also be chosen to facilitate the mapping
from the parse tree to database objects. For example, in the parse
tree for \qit{which rock contains magnesium}, shown in figure
\ref{semantic_tree}, the existence of a \sys{Specimen\_question} node
could direct the system to consult database tables containing
information about specimens, rather than tables containing information
about departures and arrivals. Similarly, the \sys{Contains\_info}
subtree could direct the system to search a table called
\sys{Contains\_info}, and locate the rows of which the
\sys{Substance} column is \qit{magnesium}.
According to \cite{Perrault}, semantic grammars are
used in {\sc Planes} \cite{Waltz}, {\sc Ladder} \cite{Hendrix2}, and
{\sc Rel} \cite{Thompson3} \cite{Thompson4}. A semantic grammar is
also used in {\sc Eufid} \cite{Templeton}.

\begin{figure}
\begin{center}
\mbox{\epsfysize=1.4in \epsffile{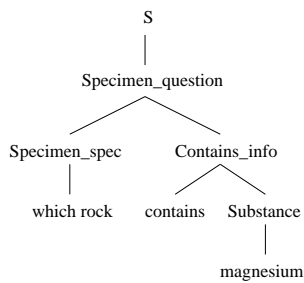}}
\caption{Parse tree in a semantic grammar}
\label{semantic_tree}
\end{center}
\end{figure}

Semantic grammars were introduced as an engineering methodology, which
allows semantic knowledge to be easily included in the system.
However, since semantic grammars contain hard-wired knowledge about a
specific knowledge domain, systems based on this approach are very
difficult to port to other knowledge domains (see section
\ref{portability}): a new semantic grammar has to be written whenever
the {\sc Nlidb} is configured for a new knowledge domain. For example,
the semantic grammar of figure \ref{sem_gram_fig} is completely
useless in an application where the database contains information
about employees and salaries.  In contrast, (at least) some of the
information of the grammar in the previous section (e.g.\ that a
sentence consists of a noun phrase followed by a verb phrase) could
probably still be used in an application where the questions refer to
employees and their salaries.

\subsection{Intermediate representation languages} \label{irls}

Most current {\sc Nlidb}s first transform the natural language question into
an intermediate logical query, expressed in some internal meaning
representation language.  The intermediate logical query expresses the
meaning of the user's question in terms of high level world concepts,
which are independent of the database structure. The logical query is
then translated to an expression in the database's query language,
and evaluated against the database. Actually, many modern natural
language front-ends (e.g.\ {\sc Cle} \cite{Alshawi}) use several
intermediate meaning representation languages, not just one.

Figure \ref{irl_arch} shows a possible architecture of an intermediate
representation language system. A similar architecture is used in {\sc
  Masque/sql} \cite{Androutsopoulos} \cite{Androutsopoulos3}.  The
natural language input is first processed syntactically by the parser.
The parser consults a set of syntax rules, and generates a parse tree,
similar to the one shown in figure \ref{lunar_tree}. (The reader is
referred to \cite{Gazdar2} for information on parsing techniques.) The
semantic interpreter subsequently transforms the parse tree to the
intermediate logic query, using semantic rules similar to the mapping
rules of section \ref{syntax_based}.

\begin{figure}
\begin{center}
\mbox{\epsfysize=3.3in \epsffile{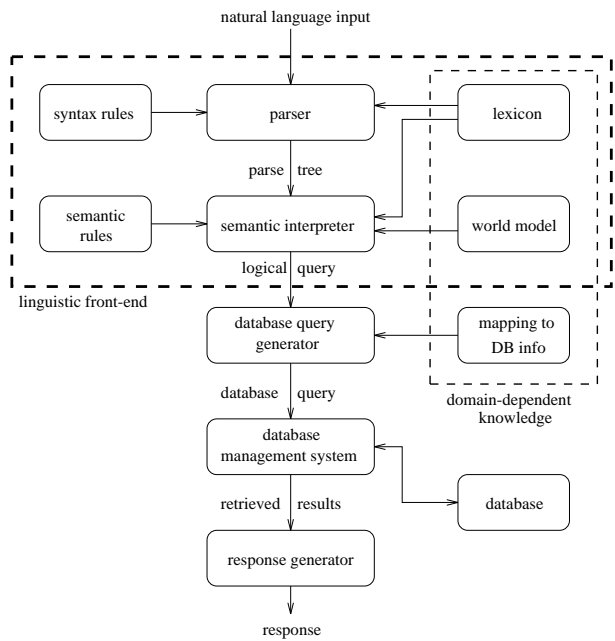}}
\caption{Possible architecture of intermediate representation language system}
\label{irl_arch}
\end{center}
\end{figure}

Some systems (e.g.\ {\sc Janus} \cite{Hinrichs1}) build on the
Montague-semantics tradition \cite{Main} \cite{Dowty}, and carry out
the semantic interpretation in a compositional, rule-to-rule manner.
Each syntax rule is coupled to a semantics rule.  The semantics rule
computes the logic expression of the constituent in the left-hand side
of the syntax rule, as a function of the logic expressions of the
constituents in the right-hand side of the syntax rule. The logic
expressions corresponding to words are declared in the lexicon.

In early {\sc Nlidb}s the syntax/semantics rules were based on rather ad hoc
ideas, and expressed in idiosyncratic formalisms.  In modern systems the
syntax/semantics rules are becoming increasingly influenced by
principled linguistic theories, and they are often expressed in
variations of well-known formalisms. {\sc Loqui} \cite{Binot}, for example,
uses a grammar influenced by {\sc Gpsg} \cite{Gazdar1}, and {\sc Cle}'s
\cite{Alshawi} grammar is expressed in a unification-based formalism
similar to {\sc Patr-II} \cite{Shieber}.

In many systems the syntax rules linking non-terminal symbols
(non-leaf nodes in the parse tree) and the corresponding semantic
rules are domain-independent; i.e. they can be used in any application
domain.  The information, however, describing the possible words (leaf
nodes) and the logic expressions corresponding to the possible words
is domain-dependent, and has to be declared in the {\em lexicon}.

In {\sc Masque} \cite{Auxerre1} \cite{Auxerre2}, for example, the lexicon
lists the possible forms of the words that may appear in the user's
questions (e.g. \qit{capital}, \qit{capitals}, \qit{border},
\qit{borders}, \qit{bordering}, \qit{bordered}\/), and logic
expressions describing the meaning of each word. For example, the
logic expression of \qit{capital}\/ (as in \qit{What is the
  capital of Italy?}\/) could be $capital\_of(Capital, Country)$,
where $Capital$ is a variable standing for a capital, and $Country$ is
a variable standing for the corresponding country. 

Similarly, the logic expression of the verb \qit{to border}\/ (as in
\qit{Which countries border Germany?}\/) could be $borders(Country_1,
Country_2)$, and the logic expression of the noun \qit{country}\/ (as
in \qit{Italy is a country}\/) could be $is\_country(Country)$.

The semantic interpretation module uses the logic expressions of the
words in the lexicon, to generate the logical query. For example, in
{\sc Masque}, the question \qit{What is the capital of each country
  bordering Greece?}\/ would be mapped to the following logical query.
(Variable names start with capital letters.)

\begin{verbatim}
answer([Capital, Country]):-
   is_country(Country), 
   borders(Country, greece), 
   capital_of(Capital, Country).
\end{verbatim}

\noindent The logic query above states that the meaning of the user's
question is to find all pairs \sys{[Capital, Country]}, such that
\sys{Country} is a country, \sys{Country} borders Greece, and
\sys{Capital} is the capital of \sys{Country}. The reader will have
noticed that the logical query was constructed using the logic
expressions of \qit{capital}, \qit{country}, and \qit{to border}.

The semantic interpreter of figure \ref{irl_arch} also consults a {\em
  world model}, that describes the structure of the surrounding
world.  Typically, the world model contains a {\em hierarchy} of
classes of world objects, and {\em constraints}\/ on the types of
arguments each logic predicate may have.
Figure \ref{hierarchy} shows a world-model hierarchy of {\sc Masque},
for a business application. Similar hierarchies are used in {\sc Team}
\cite{Grosz1} \cite{Grosz2} and {\sc Cle} \cite{Alshawi}.  The
hierarchy of figure \ref{hierarchy} states that, in the particular
application domain, a person can be either a client or an employee,
that employees are divided into technicians, salespersons, and
managers, etc.
The hierarchy is typically used in combination with constraints on
predicate arguments. These constraints specify the classes to which
the arguments of a logic predicate may belong.

\begin{figure}
\begin{center}
\mbox{\epsfxsize=2.8in \epsffile{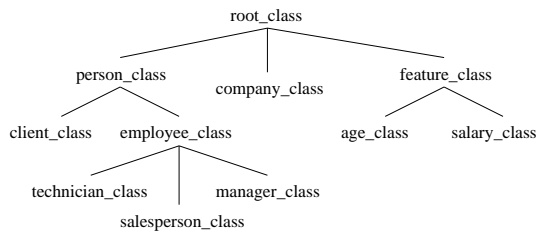}}
\caption{A hierarchy in a world model}
\label{hierarchy}
\end{center}
\end{figure}

To illustrate the need for predicate-argument constraints, let us
consider the business application of figure \ref{hierarchy}, and let
us assume that the underlying database holds information about the
salaries of employees, but that it does not hold information about the
salaries of people not employed by the company. (Let us also assume
that an employee cannot also be a client of the company.) Then, the
question \qit{What is the salary of each client?}\/ will retrieve no
information, because the database does not contain information about
the salaries of clients.  We would like the {\sc Nlidb} to be able to detect
that the question is conceptually problematic.

Let us assume that the logic expression of \qit{salary}\/ is
$salary\_of(Salary, Person)$, and that the logic expression of
\qit{client}\/ is $is\_client(Person)$. The corresponding {\sc Masque}
logical query would be:

\begin{verbatim}
answer([Salary, Person]):- 
   is_client(Person), 
   salary_of(Salary, Person).
\end{verbatim}

Using predicate-argument constraints, we could specify that the argument
of $is\_client$ must belong to the class (must be of type) $client\_class$, and
that the arguments of $salary\_of$\/ must belong to the classes $salary\_class$
and $employee\_class$ respectively. These constraints would require the
variable \sys{Person} in the logical query to belong to both
$client\_class$ and $employee\_class$. Since, neither $client\_class$ nor
$employee\_class$ is a descendant of each other in the hierarchy of figure
\ref{hierarchy}, the system would be able to reason that the logical query
contains a class mismatch. The class mismatch could then be reported
to the user. 
In contrast, assuming that the logic expression of \qit{technician}\/ 
is $is\_technician(Person)$, where $Person$ is a variable of type
$technician\_class$, the question \qit{What is the salary of each technician?}
will generate the {\sc Masque} logical query:

\begin{verbatim}
answer([Salary, Person]):- 
   is_technician(Person), 
   salary_of(Salary, Person).
\end{verbatim}

\noindent In this case, $Person$ is required to belong to
$technician\_class$ {\em and}\/ $employee\_class$. Since
$employee\_class$ is an ancestor of (i.e.\ subsumes)
$technician\_class$, the logical query contains no class mismatch.

The predicate-argument constraints can also help the system to resolve
natural language ambiguities. Consider, for example, the request
\qit{List all employees in a company with a driving licence}. From the
{\sc Nlidb}'s point of view, the request contains a modifier attachment
ambiguity (see section \ref{modif_att}): \qit{with a driving licence}
may refer either to \qit{employees} or to \qit{company}. In the first
case, a {\sc Masque}-like logic query would be:

\begin{verbatim}
answer([Employee]):- 
   is_employee(Employee),
   is_company(Company),
   is_drv_licence(Driv_lic),
   in(Employee, Company),
   has(Employee, Driv_lic).
\end{verbatim}

\noindent while in the second case (\qit{with a driving licence}\/ refers to
\qit{company}), the logic query would look like:

\begin{verbatim}
answer([Employee]):- 
   is_employee(Employee),
   is_company(Company),
   is_drv_licence(Driv_lic),
   in(Employee, Company),
   has(Company, Driv_lic).
\end{verbatim}

The world model could specify that the arguments of $is\_employee$,
$is\_company$, and $is\_drv\_licence$\/ are of types
$employee\_class$, $company\_class$, and $drv\_licence\_class$\/
respectively, and that if the second argument of $has$\/ is of type
$drv\_licence\_class$, then the first argument must be of type
$person\_class$\/ (i.e.\ only persons are allowed to have driving
licences). The hierarchy could then be used to state that
$person\_class$\/ subsumes $employee\_class$\/ but not
$company\_class$\/ (i.e.\ companies are not persons). This would rule
out the second logical query above, where the second argument of
$has$\/ is of type $drv\_licence\_class$, and the first one is not of
type $person\_class$.

\medskip

The logic query generated by the parsing and semantic interpretation
module, expresses the meaning of the user's question in terms of
logical, high-level concepts. The logic query does not refer to
database objects (e.g.\ tables, columns), and it does not specify
how to search the database to retrieve the necessary information. In
order to retrieve the information requested by the user, the logic
query has to be transformed into a query expressed in some database
query language supported by the underlying {\sc Dbms} (Database Management
System). In figure \ref{irl_arch} this transformation is carried out
by the {\em database query generator}, using the {\em mapping to
  database information}.

The mapping to database information specifies how logic predicates
relate to database objects. In the case of an interface to a relational 
database, the simplest approach would be to link each logic predicate to
an {\sc Sql} \sys{SELECT} statement, selecting rows from one or more database
tables.
Let us assume, for example, that the database contains the tables:

{\small
\begin{center}
\begin{tabular}{lr}
\begin{tabular}[t]{lll}
\multicolumn{3}{c}{countries\_table} \\
country & capital & language \\ 
\hline
France & Paris & French \\
Spain & Madrid & Spanish \\
Germany & Berlin & German \\
\dots & \dots & \dots 
\end{tabular}
&
\begin{tabular}[t]{ll}
\multicolumn{2}{l}{borders\_table} \\
country1 & country2 \\
\hline
France & Spain \\
France & Germany  \\
Germany & France  \\
\dots & \dots 
\end{tabular}
\end{tabular}
\end{center}
}

\noindent Let us also consider the {\sc Masque} logic query for
\qit{What is the capital of each country bordering Greece?}:

\begin{verbatim}
answer([Capital, Country]):-
   is_country(Country),
   borders(Country, greece),
   capital_of(Capital, Country).
\end{verbatim}

\noindent The mapping information could link the predicate
\sys{is\_country} to the {\sc Sql} query:
\begin{verbatim}
SELECT country
FROM countries_table
\end{verbatim}

\noindent Such linking would mean that \sys{is\_country} is true if its
argument appears in the column $country$\/ of $countries\_table$.
Similarly, \sys{borders} could be linked to the {\sc Sql} query:

\begin{verbatim}
SELECT country1, country2
FROM borders_table
\end{verbatim}

\noindent which would mean that \sys{borders} is true if there is a row
in $borders\_table$, such the first argument of \sys{borders} is the
$country1$\/ of the row, and the second argument of \sys{borders} is
the $country2$\/ of the row.  Finally, \sys{capital\_of} could be
linked to:

\begin{verbatim}
SELECT capital, country
FROM countries_table
\end{verbatim}

A similar (but more complex) approach is used in {\sc Masque/sql}
\cite{Androutsopoulos} \cite{Androutsopoulos3}.  Using the mapping
information, {\sc Masque/sql} would generate an {\sc Sql} query similar to:

\begin{verbatim}
SELECT countries_table.capital, countries_table.country
  FROM countries_table, borders_table
 WHERE countries_table.country = borders_table.country1
   AND borders_table.country2 = 'Greece'
\end{verbatim}

The transformation of the logic query to a database query does not
need to be made in one stage. In \cite{DeRoeck1} \cite{DeRoeck2}
\cite{Lowden1}, for example, the writers describe a principled
multi-stage transformation process, used in the {\sc Nlidb} developed
at the University of Essex. The Essex system first generates a logic
query, expressed in a version of untyped $\lambda$-calculus
\cite{Dowty}. The $\lambda$-calculus expression is then transformed
into a first-order predicate logic expression, which is subsequently
translated into universal-domain relational calculus, domain
relational calculus, tuple relational calculus, and finally {\sc Sql}
(see \cite{Ullman} for a description of the various forms of
relational calculus).

In figure \ref{irl_arch} the database query generated by the database
query generator is passed to the underlying {\sc Dbms}, which executes the
query against the database, and passes the retrieved data to the {\em
  response generator}. The latter is responsible for reporting the
retrieved information to the user.  Issues related to response
generation are discussed in the following section.

In most modern {\sc Nlidb}s the underlying database is assumed
to be relational. Some {\sc Nlidb}s support multiple underlying,
possibly heterogeneous, databases (see section
\ref{multiple_underlying}).  Research systems often assume
idiosyncratic database models, especially designed to facilitate the
development of the {\sc Nlidb}.

When constructing a new {\sc Nlidb}, one design issue is the `division
of labour' between the {\sc Nlidb} software and the {\sc Dbms}'s query
language. The {\em de facto} industry standard query language is {\sc
  Sql}.  As a language, standard {\sc Sql} is less expressive than a
programming language, since it lacks constructs for iteration and
recursion, amongst other things.  Thus there are computationally
reasonable queries that might be expressed in natural language, but
which cannot be answered by a single {\sc Sql} query. Of course, most
current {\sc Nlidb}s support only a limited subset of natural
language. In many cases, this subset may be so narrow that less
queries can be expressed in the supported natural language subset than
in {\sc Sql}. The point being made here, however, is that {\em full}\/
natural language is more expressive than single {\sc Sql} queries.
Thus, as the natural language subset supported by the linguistic component
grows, there will be inevitably natural language queries that cannot
be mapped to single {\sc Sql} queries.

In such cases, the designer of the {\sc Nlidb} must choose between
either restricting the natural language subset to make it match the
expressivity of single {\sc Sql} queries, or allowing a natural
language query to be translated into some data-dependent number of
{\sc Sql} queries, and piecing together the answer to the original
natural language query from the set of answers to the {\sc Sql}
queries.  Thus the iteration or recursion would have to be controlled
from within the {\sc Nlidb} software.

Developments in {\sc Sql} are likely to have an impact on the design
of {\sc Nlidb}s.  {\sc Sql3} has a richer set of storage structures,
types and querying constructs than {\sc Sql}.  On the one hand, this
means that more can be done in the {\sc Dbms}, and less in the {\sc
  Nlidb}, but on the other hand, broader linguistic coverage will be
needed in the {\sc Nlidb} to ensure that the supported natural
language subset captures the expressivity of the {\sc Dbms}'s query
language.

\medskip

One of the advantages of {\sc Nlidb}s based on intermediate
representation languages is that the {\em linguistic front-end}\/ (the
part of the system that generates the logic queries -- see figure
\ref{irl_arch}) is independent of the underlying {\sc Dbms}. Thus, the
{\sc Nlidb} can be ported to a different {\sc Dbms}, by rewriting the
database query generator module (see section \ref{dbms_portability}).
This approach also allows the development of generic linguistic
front-ends, that can be used as parts of interfaces to systems other
than databases (e.g. expert systems, operating systems etc).  One
example of such a generic front-end is the {\sc Cle} system
\cite{Alshawi}.

Another advantage of the particular architecture shown in figure
\ref{irl_arch}, is that the domain-dependent knowledge (i.e. knowledge
that has to be modified whenever the system is used in a new knowledge
domain -- e.g. questions about flights, instead of questions about books
in a library) is clearly separated from the rest of the front-end, thus
allowing knowledge-domain portability (see section
\ref{knowledge_domain_portability}).

Finally, {\sc Nlidb}s based on logical intermediate representation languages,
allow reasoning modules to be added between the semantic interpreter
and the database query generator, so that the {\sc Nlidb} can carry out
reasoning based on the information stored in the database (see section
\ref{intelligent}). In systems where the parse tree is mapped directly
to an expression in a database query language (e.g.\ {\sc Sql}), such
reasoning would have to be carried out in the database query language.
This is practically impossible, since database query languages are not
designed to facilitate machine reasoning.

\subsection{Response generation} \label{response_generation}

The discussion in section \ref{irls} focused on the interpretation of
the natural language input. Generating a suitable response is also
important. Some {\sc Nlidb}s that interface to relational databases
simply print the tuples retrieved by the database query, an approach
which is not always acceptable:

\begin{itemize}

\item The retrieved database tuples may contain encoded information (e.g.\
department codes instead of department names).

\item The system may not be able to ``understand'' the question, in which
  case no tuples will be retrieved. The cause of the failure should
  be explained to the user (e.g.\ unknown word, syntax too complex, no
  relevant information in the database, etc).

\item The user's question may contain false presuppositions about the
database or the world, in which case the system should report the false
presuppositions (e.g.\ \qit{Do all managers earning more than 
  50,000 work in sales?}, when no manager earns more than 50,000).

\item The user's request may not express literally what the user wants
  to know. For example, it may not be acceptable to answer a
  \qit{yes}/\qit{no} question (e.g.\ \qit{Is there a flight to
    Athens?}\/) by a yes/no response.

\end{itemize}

To illustrate some of these points, consider the following dialogue
between the user and {\sc Loqui}, based on examples from \cite{Loqui} and
\cite{Binot}:

\smallskip
\begin{tabular}{l}
\sys{>} \usr{Does every person that works on 20 projects work on {\sc
    Hoscom}?} \\ 
\sys{There is no such person.} \\
\sys{>} \usr{Does David Sedlock work on something?} \\
\sys{Yes.} \\
\sys{{\sc Bim}\_{\sc Loqui}, {\sc Mmi2}, {\sc Nlpad}} \\
\sys{>} \usr{Does RV work on {\sc Loqui}?} \\
\sys{No. RV works on {\sc Bim-Prolog}.}
\end{tabular}
\smallskip

In the first question, the system has detected the false
presupposition that there are people working on 20 projects, and it
has reported a suitable message. (Strictly logically speaking the
response could have been \qsys{yes}, since no person works on 20
projects.)  In the second and third questions, the system did not
simply print \qit{yes}/\qit{no} responses; it also printed additional
information, that the user probably wanted to know.

In some cases, such ``cooperative responses'' (see \cite{Kaplan}
\cite{Kaplan2}) can be generated by using relatively simple
mechanisms.  For example, in yes/no questions like \qit{Does David
  Sedlock work on something?}, if the answer is affirmative one
strategy is to print a ``\sys{yes}'', followed by the answer to
\qit{On what does David Sedlock work?}. (It is usually not difficult
to generate automatically questions like \qit{On what does David
  Sedlock work?} from questions like \qit{Does David Sedlock work on
  something?}.) A description of some relatively simple mechanisms
that have been used in {\sc Masque} to generate cooperative responses
can be found in \cite{Sentance}, and \cite{Gal1991} gives a more
formal account of some of these techniques.

It is not always possible, however, to generate reasonable cooperative
responses by using simple mechanisms. In some cases, in order to
generate an acceptable cooperative response, the {\sc Nlidb} must be able to
reason about the user's goals. This is discussed in section
\ref{user_model}.

Another problem is that the {\sc Nlidb} may misinterpret a question
submitted by the user, without the user becoming aware that his/her
query has been misinterpreted. Natural language questions often have
several readings; the {\sc Nlidb} may select a reading of a question
that is different from the reading the user had in mind when he/she
typed the question. In these cases, it may be hard for the user to
understand that the system has actually answered a different question.
To avoid such misunderstandings, {\sc Tqa} \cite{Damerau} contains a
module that converts the {\sc Sql} query back to natural language. The
reconstructed natural language request is presented to the user, to
ensure that none of the intermediate transformation stages has caused
his/her request to be misinterpreted. Similar paraphrasing modules are
used in several other {\sc Nlidb}s. \cite{DeRoeck3} \cite{Lowden3}
describes how the paraphrasing module of a particular {\sc Nlidb}
works.

To make the retrieved information easier to read, many {\sc Nlidb}s
contain response formatting modules, that allow encoded
information retrieved from the database (e.g.\ \qsys{dpt341}) to be
expanded to more natural formats (e.g.\ \qsys{Sales department}).
Finally, some {\sc Nlidb}s (e.g.\ {\sc Natural Language}
\cite{Natural_Language}) can present some kinds of retrieved data in
graphics (e.g.\ pie-charts).

\subsection{Multiple underlying systems} \label{multiple_underlying}

The architectures described so far, allow the user to interact with a single
database. In some cases, however, access to more than one
underlying applications may be necessary.

In certain domains, the information needed to answer the user's
question may be distributed in several, possibly heterogeneous,
databases. (See \cite{Ceri3} for an introduction to
distributed and heterogeneous databases.) Some {\sc Nlidb}s (e.g.\ 
{\sc Intellect} according to \cite{Intellect}) allow the user to
access transparently multiple databases through a single English
question.  The {\sc Nlidb} translates the logic query into a set of
database queries, each one expressed in the query language of the
corresponding {\sc Dbms}.  The {\sc Nlidb} then collects and joins the
partial results retrieved by each {\sc Dbms}, and reports the answer
to the user.

{\sc Intellect} can be configured to interact with the database
through the {\sc Kbms} \cite{KBMS} expert system shell. This way a
reasoning layer can be added between the {\sc Nlidb} and the database,
which allows the {\sc Nlidb} to carry out reasoning based on the data
stored in the database (see section \ref{intelligent}).

{\sc Ask} \cite{Thompson1} \cite{Thompson2} can be configured to
interface transparently to several kinds of software systems (e.g.
electronic mail systems, screen painters, word processors, etc). {\sc
  Ask} translates each natural language request into a suitable
expression, which is then directed to the appropriate underlying
system. This way, the user views an integrated computer system, with
natural language understanding abilities.

{\sc Janus} \cite{Bobrow2} has the ability to interact with databases,
expert systems, decision support systems, graphics systems, etc., and
all the underlying systems may be involved in the answering of a
user's question. For example (the example is based on \cite{Bobrow2}),
when asked \qit{Which submarines have the greatest probability of
  locating A within 10 hours?}, the system might use a database to
retrieve the current positions of the submarines, and two separate
expert systems to compute (a) the navigation course each submarine would
have to follow, and (b) the probability of success for each submarine.

\section{Portability} \label{portability}

Early {\sc Nlidb}s were each designed for a particular database
application.  {\sc Lunar} \cite{Woods}, for example, was designed to
support English questions, referring to a database of a particular
structure, holding data about moon rocks. Such application-tailored
{\sc Nlidb}s were very difficult to port to different applications.
In recent years a large part of the research in {\sc Nlidb}s has been
devoted to portability, i.e. to the design of {\sc Nlidb}s that can be
used in different knowledge-domains, with different underlying {\sc
  Dbms}s, or even with different natural languages. This section
discusses the different kinds of {\sc Nlidb} portability, and presents
some methods that have been used to approach the goal of portability.
Most of the discussion relates to the architecture of figure
\ref{irl_arch}.

\subsection{Knowledge-domain portability} \label{knowledge_domain_portability}

Existing {\sc Nlidb}s can only cope with questions referring to a particular
{\em knowledge domain}\/ (e.g.  questions about train services,
questions about bank accounts). A {\sc Nlidb} provides knowledge-domain
portability, if it can be configured for use in a wide variety of
knowledge domains.

Typically, whenever a {\sc Nlidb} is being reconfigured for a new
knowledge-domain, someone has to ``teach'' the system the words and
concepts used in the new domain, and how these concepts relate to the
information stored in the database. In systems adopting architectures
similar to the one of figure \ref{irl_arch}, this ``teaching''
modifies the lexicon, the world model, and the mapping to the
database.

Different {\sc Nlidb}s assume that the knowledge-domain configuration will be 
carried out by people possessing different skills:

\paragraph{Programmer:} In some systems a part (usually small and well
  defined) of the {\sc Nlidb}'s code has to be rewritten during the
  knowledge domain configuration. It is, therefore, assumed that the person
  that will carry out the configuration will be a programmer,
  preferably familiar with the {\sc Nlidb}'s code.

\paragraph{Knowledge engineer:} Other {\sc Nlidb}s provide tools that
can be used to configure the system for a new knowledge domain,
without the need to do any programming. It is still assumed, however,
that the tools will be used by a knowledge engineer, or at least a
person familiar with basic knowledge representation techniques,
databases, and linguistic concepts.

In the following dialogue, between {\sc Masque}'s ``domain editor''
\cite{Auxerre2} and the knowledge engineer, the verb \qit{to exceed}\/
(as in \qit{The population of Germany exceeds the population of
  Portugal.}\/) is added to the system's knowledge.  (Actually, {\sc
  Masque}'s domain-editor does not shield completely the
knowledge-engineer from the programming language.  The knowledge
engineer is still required to write some Prolog code to map logic
predicates to database objects.)

\smallskip
\noindent
\verb+editor>+ {\em add verb} \\ 
\verb+what is your verb ?+ {\em exceed} \\ 
\verb+what is its third sing. pres ?+ {\em exceeds} \\ 
\verb+what is its past form ?+ {\em exceeded} \\ 
\verb+what is its perfect form ?+ {\em exceeded} \\ 
\verb+what is its participle form ?+ {\em exceeding} \\ 
\verb+to what set does the subject belong ?+ {\em numeric} \\ 
\verb+is there a direct object ?+ {\em yes} \\ 
\verb+to what set does it belong ?+ {\em numeric} \\ 
\verb+is there an indirect object ?+ {\em no} \\ 
\verb+is it linked to a complement ?+ {\em no} \\ 
\verb+what is its predicate ?+ {\em greater\_than} \\ 
\verb+do you really wish to add this verb?+ {\em y}
\smallskip

\noindent The logic expression of \qit{to exceed}\/ was defined to be
$greater\_than(arg_1,arg_2)$, where both arguments were declared to
belong to the hierarchy class $numeric$ (see section \ref{irls}).

Some systems (e.g. {\sc Parlance} \cite{BBN}, {\sc Cle} \cite{Alshawi}) provide
morphology modules, that allow the system to determine the various
forms of the words using stems stored in the lexicon and morphology
rules.  This way, the user does not have to specify all the word forms
as in the example above (\qit{exceed}, \qit{exceeds}, \qit{exceeded},
\qit{exceeding}, etc). This ability is particularly important in
highly-inflected languages, where each word may have numerous possible
forms. Some systems (e.g. {\sc Natural Language} \cite{Natural_Language},
{\sc Cle} \cite{Alshawi}) have built-in dictionaries, listing the most
common words. The dictionaries can be customised by the person
configuring the system to include domain-specific terminology.

\paragraph{DB administrator:} According to the designers of {\sc Team}
  \cite{Grosz1} \cite{Grosz2}, a realistic scenario is to assume that
  the {\sc Nlidb} will be bought by a company already using a database to
  which the {\sc Nlidb} is to be linked.  In such situations, the
  responsibility of installing and configuring the {\sc Nlidb} would
  probably be assigned to the local database administrator.
  Therefore, the configuration tools should be designed, so that they
  can be used by persons with a good understanding of database
  concepts, familiar with the particular database to which the {\sc Nlidb}
  is to be linked, but with no knowledge of AI, logics, or
  linguistics. {\sc Team} tries to infer what questions can be asked about
  the database, by eliciting from the database administrator
  information about the use and structure of the database.

  In the following dialogue (from \cite{Grosz1} -- slightly
  simplified), the system collects information about the structure and
  use of the database table (``\verb+file+'') \sys{CHIP}. (The reader
  is reminded that messages generated by the system are shown in
  \sys{this typeface}, and user entries are shown in \usr{this
    typeface}.)

\smallskip
\begin{tabular}{l}
\sys{file name: CHIP}\\
\sys{fields:} \usr{ID MAKER WIDTH} \dots \\
\sys{subject:} \usr{processor} \\
\sys{synonyms for processor:} \usr{chip} \\
\sys{primary key:} \usr{ID} \\
\sys{Can one say ``Who are the processors?':} \usr{no} \\
\sys{Pronouns for subject (he, she, it,they):} \usr{it} \\
\sys{field: MAKER} \\
\sys{type of field (symbolic, arithmetic, feature):} \usr{symbolic} \\
\dots
\end{tabular}
}
\smallskip

The user is asked to provide information about the table's fields
(attributes), the subjects the table describes (\qit{processor}), etc.
The dialogue would then continue with the system asking several
questions for each field of \sys{CHIP}. The reader will have noticed
that the system's questions do not refer to logic concepts (e.g.
hierarchy classes, predicates) or advanced linguistic concepts,
although they contain database terminology (e.g. \qsys{primary key}).

\paragraph{End-user:} Other {\sc Nlidb} designers emphasise that the
configuration phase can never be complete, and that the {\sc Nlidb}
will always have to learn new words and concepts.  Therefore, the {\sc
  Nlidb} should allow the end-users to teach the system on-line, and
using definitions expressed in natural language.  Systems adopting
this approach usually have a large built-in vocabulary, and a set of
primitive concepts (meanings). The end-user can extend the system's
capabilities by teaching it new words, and by defining the meanings of
the new words in terms of primitive concepts or in terms of words the
system already knows.

The following examples from \cite{BBN} show some of the capabilities
of {\sc Parlance}:

\smallskip
\begin{tabular}{l}
\sys{>} \usr{define: a yuppie is a person under 30 with a graduate
  degree who}\\
\sys{ } \usr{earns over 5K per month}\\
\sys{>} \usr{define: a rich employee is an employee who earns over
  100K per year}  
\end{tabular}

\subsection{DBMS portability} \label{dbms_portability}

A {\sc Nlidb} provides {\sc Dbms} portability, if it can be easily
modified to be used with different underlying database management
systems ({\sc Dbms}s).

In the case of {\sc Nlidb}s that generate queries in widely-supported
database query languages (e.g.\ {\sc Sql}), it may be possible to port
the {\sc Nlidb} to another {\sc Dbms}, with only minor modifications,
if the new {\sc Dbms} supports the same database query language as the
old one.

If the original database query language is not supported by the new
{\sc Dbms}, then a {\sc Nlidb} based on the architecture of figure
\ref{irl_arch} can be ported to the new {\sc Dbms} by rewriting the
module that translates the intermediate logic queries to database
queries. In contrast, {\sc Nlidb}s that do not use intermediate
representation languages may require extensive modifications before
they can be used with a new {\sc Dbms} that does not support the old
database query language.

In the case of {\sc Nlidb}s that clearly separate the linguistic
front-end (see figure \ref{irl_arch}) from the rest of the system, it
should be possible to port a {\sc Nlidb} configured for a particular
knowledge-domain to a different underlying {\sc Dbms} containing data
about the same knowledge-domain, without modifying the modules of the
linguistic front-end.

\subsection{Natural language portability} \label{nl_portability}

The largest part of the research that has been carried out in the area
of {\sc Nlidb}s assumes that the natural language requests will be
written in English.  Modifying an existing {\sc Nlidb} to be used with
a different natural language can be a difficult task, since the {\sc
  Nlidb} may contain deeply embedded assumptions about the natural
language to be used.

In {\sc Nlidb}s based on the architecture of figure \ref{irl_arch},
modifying the {\sc Nlidb} for a different natural language typically
requires rewriting/modifying the syntax rules, the semantic rules,
and the lexicon.
An attempt to build a Portuguese version of {\sc Chat-80} \cite{Warren} is
described in \cite{Lopes}. Information about a Swedish version of {\sc Cle}
can be found in \cite{Alshawi}.

\subsection{Hardware and programming language portability}

Up to the early eighties, the ability to port the {\sc Nlidb} to
different hardware and software environments were also important
aspects of {\sc Nlidb}s. Some {\sc Nlidb}s could be used only on
expensive research computing systems, supporting ``exotic''
programming languages (e.g.\ Lisp machines), thus making {\sc Nlidb}s
almost impossible to use in real-life applications.  In recent years
the availability of powerful low-cost computers and the fact that
``AI'' languages like Prolog and Lisp are now commonly available have
allowed {\sc Nlidb}s to become easy to port across computing
platforms.

\section{Restricted NL input} \label{restricted}

Current {\sc Nlidb}s support only a limited subset of natural language. The
limits of this subset are usually not obvious to the user (see section
\ref{comparison}). Users are often forced to rephrase their questions,
until a form the system can understand is reached. In other cases,
users never try questions the {\sc Nlidb} could in principle handle, because
they think the questions are outside the subset of natural language
supported by the {\sc Nlidb}. 

Many research projects attempt to expand the linguistic coverage of
{\sc Nlidb}s, hoping that eventually {\sc Nlidb}s will be able to
understand most user requests, at least in particular knowledge
domains. An alternative approach, presented in this section, is to
deliberately and explicitly restrict the set of natural language
expressions the user is allowed to input, so that the limits of this
subset are obvious to the user. By choosing carefully the subset of
the allowed natural language expressions, the design and
implementation of the {\sc Nlidb} can also be simplified.

This section presents two approaches that drastically restrict the set
of natural language expressions the user is allowed to input. In the
first one, only questions of a very specific syntactic pattern are
allowed; in the second one, the user has to compose his/her queries by
choosing words from limited lists of words displayed on the screen.

\subsection{Restricted NL syntax}
\label{restricted_syntax}

In {\sc Pre} \cite{Epstein} the user is allowed to input only
questions of a specific syntactic pattern. All questions must have the
form:

\begin{center}
\begin{tabular}{l}
{\em what}\\
{\em conjoined noun phrases} \\
\ \ \ {\em nested relative clauses} \\ 
\ \ \ \ \ \ {\em conjoined relative clauses}
\end{tabular}
\end{center}

\noindent The following question (from \cite{Epstein} -- simplified) is an
example of a question acceptable by {\sc Pre}:
\begin{center}
\begin{tabular}{ll}
\usr{what are}\\
\usr{the names, ids, and categories of the employees} & (1)\\
\ \ \ \usr{who are assigned schedules} & (2) \\
\ \ \ \usr{that include appointments} & (3) \\
\ \ \ \usr{that are executions of orders} & (4) \\
\ \ \ \ \ \ \usr{whose addresses contain `maple' and} & (5) \\
\ \ \ \ \ \ \usr{whose dates are later than 12/15/83 and} & (6) \\
\ \ \ \ \ \ \usr{whose statuses are other than `comp'} & (7)
\end{tabular}
\end{center}

In the example above, line (1) contains the conjoined noun phrases. Lines
(2)--(4) contain the nested relative clauses. Notice that these clauses are
nested, because (3) refers to \qit{schedules}\/ in (2), and (4) refers to
\qit{appointments}\/ in (3). Lines (5)--(7) contain the conjoined
relative clauses. The clauses in (5)--(7) are not nested; they all refer
to \qit{orders}\/ in (4).

\noindent In contrast, the following question is not acceptable by {\sc Pre}:
\begin{center}
\begin{tabular}{ll}
\usr{what are} \\
\usr{the addresses of the appointments} & (1) \\
\ \ \ \usr{that are included in schedules} & (2) \\
\ \ \ \ \ \ \usr{whose call times are before 11:30 and} & (3) \\
\ \ \ \ \ \ \usr{that are executions of orders} & (4) \\
\ \ \ \ \ \ \ \ \ \usr{whose statuses are other than `comp'} & (5)
\end{tabular}
\end{center}

In the latter example, line (2) contains the only clause of the {\em nested 
relative clauses}\/ part. Both (3) and (4) are conjoined relative clauses
referring to \qit{schedules}\/ in (2). However, (5) is nested with respect to
(4), since it refers to \qit{orders}\/ in (4), not to \qit{schedules}\/ in
(2). The {\sc Pre} question pattern does not allow the {\em conjoined relative 
clauses}\/ part to be followed by another {\em nested relative clauses}\/
part, and hence the question is ill-formed.

\begin{figure}
\begin{center}
\mbox{\epsfysize=2in \epsffile{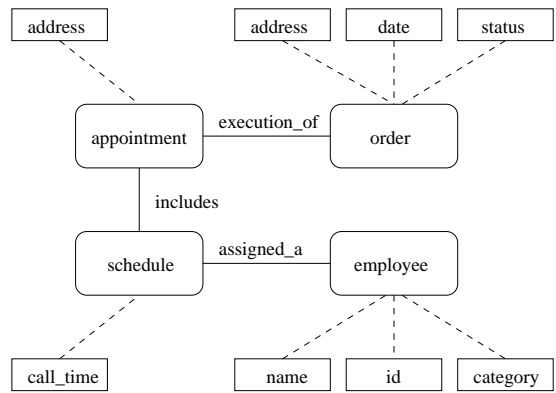}}
\caption{A {\sc Pre}-like database schema}
\label{PRE_graph}
\end{center}
\end{figure}

\cite{Epstein} claims that the {\sc Pre} question pattern allows
the users to have a clear view of the questions the system is able to
answer, and that the pattern is easy to remember.
The {\sc Pre} question pattern is designed to facilitate the mapping from
user questions to navigation graphs for {\sc Pre}'s entity-relationship-like
\cite{Ullman} database model. Figure \ref{PRE_graph} shows the
database schema to which the above two examples refer. (The schema of
figure \ref{PRE_graph} may not have exactly the same form as the
schemata used in {\sc Pre}.)  Oval boxes correspond to entity classes,
rectangular boxes correspond to attributes, and continuous lines
correspond to relations.

\begin{itemize}

\item The {\em conjoined noun phrases}\/ part of each {\sc Pre} question
corresponds to a projection operator, similar to the ones used in relational 
algebra \cite{Ullman}.
Projection operators determine the attributes of an entity to be reported. 

\item The {\em nested relative clauses}\/ part corresponds to traversals of
continuous line links in the {\sc Pre} model.

\item The {\em conjoined relative clauses}\/ part corresponds to 
select operators, similar to the ones used in relational algebra.
A select operator selects entities from an entity class, 
according to some criteria.

\end{itemize}

To answer the question of the first example in this section, {\sc Pre} first
transforms the {\em conjoined relative clauses}\/ part of the question
to a select operator, that selects all entities from class
\sys{order}, containing \qit{maple}\/ in their \sys{address}es, having
\sys{date}s later than 12/15/83, and having \sys{status}es other than
\qit{comp}.

Next, {\sc Pre} transforms the {\em nested relative clauses}\/ part of the
question to a sequence of relation (continuous line) traversals. In the
case of our example, the nested relative clauses are mapped to a 
sequence of traversals, mapping \sys{order}s to \sys{appointment}s via the
\sys{execution\_of} relation, \sys{appointment}s to \sys{schedule}s
via the \sys{includes} relation, and \sys{schedule}s to 
\sys{employee}s via the \sys{assigned\_a} relation.

Finally, {\sc Pre} transforms the {\em conjoined noun phrases}\/ part of the 
question to a projection operator. For each \sys{employee} entity reached
during the previous step, the system reports its \sys{name}, \sys{id}, and
\sys{category}.

\subsection{Menu-based systems} \label{menu-based} 

In menu-based systems the user is not able to type directly his/her
questions.  Instead, each question has to be constructed by choosing
possible words or phrases from menus. The menus are displayed on the
screen, and they are continuously updated.  This method, used in the
{\sc Nlmenu} system \cite{Tennant1} \cite{Tennant2}, will be
illustrated by examples from \cite{Tennant2}. Q\&A (section
\ref{history}) provides a similar menu-based interface.

\begin{figure}[t]
\begin{center}
\mbox{\epsfysize=2.9in \epsffile{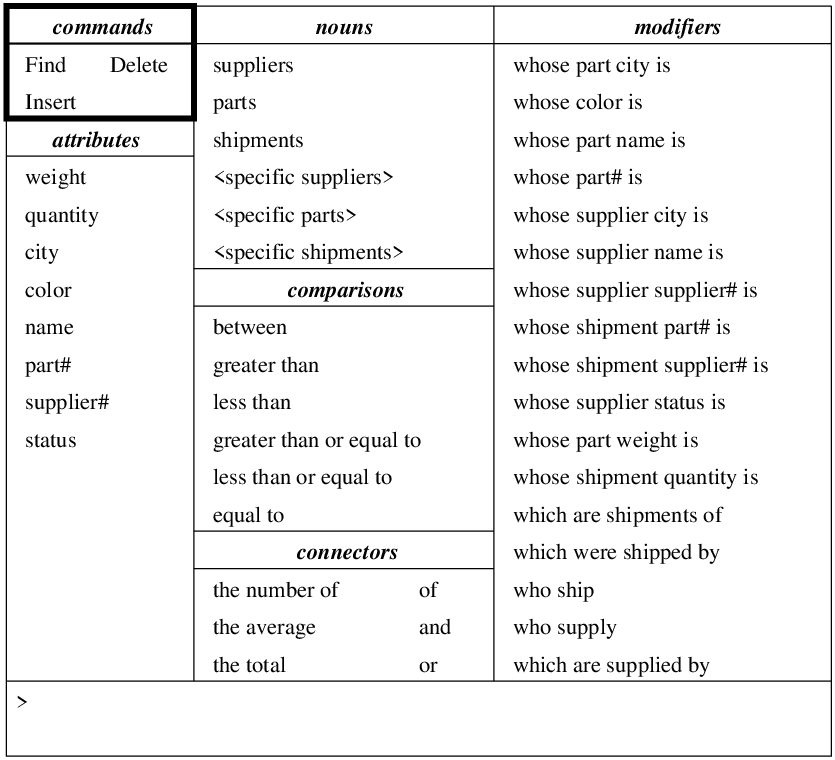}}
\caption{The initial screen of {\sc Nlmenu}}
\label{nlmenu2}
\end{center}
\end{figure}

\begin{figure}[t]
\begin{center}
\mbox{\epsfysize=2.9in \epsffile{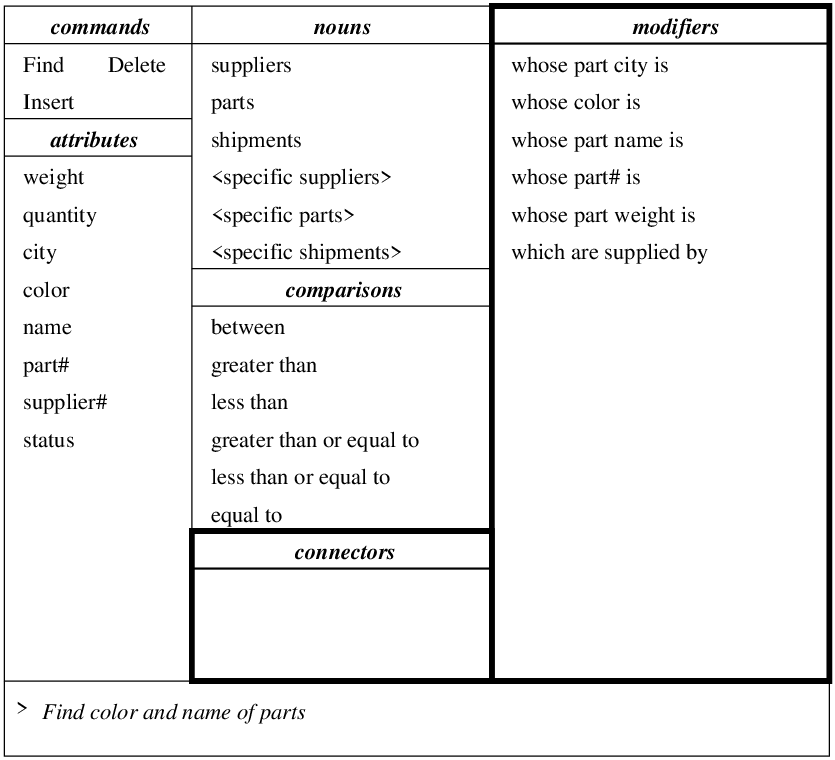}}
\caption{The {\sc Nlmenu} screen after several more words have been selected}
\label{nlmenu4}
\end{center}
\end{figure}

{\sc Nlmenu} first displays a screen like the one shown in figure
\ref{nlmenu2}. (The actual {\sc Nlmenu} screens are slightly more complicated.)
Highlighted borders indicate menus that are currently
active.  The user first selects \qit{Find}. ({\sc Nlmenu} also has update
capabilities -- see section \ref{updates} -- which will not be
discussed in this paper.)  \qit{Find}\/ becomes the first word of the
user's question, and it is displayed in the lower part of the screen.
After \qit{Find}\/ has been selected, the \sys{attributes},
\sys{nouns}, and \sys{connectors} menus become activated. The user
selects \qit{color} from the \sys{attributes} menu, and the question
becomes \qit{Find color}. In a similar manner, the user then selects
\qit{and}, \qit{name}, \qit{of}, and \qit{parts}. At this stage the
{\sc Nlmenu} screen has become as shown in figure \ref{nlmenu4}. (After
selecting each word, the contents of the menus are modified.)

The user selects \qit{whose color is}\/ from the \sys{modifiers} menu,
and the question becomes \qit{Find color and name of parts whose color
  is}. At this point, a special menu appears on screen, listing the
possible colors.  After selecting a color (e.g.\ \qit{red}\/), the
question is complete. The user executes the query by selecting an
option on the screen, and the system reports the results retrieved
from the database.

\medskip

{\sc Nlmenu} uses a context-free semantic grammar (see section
\ref{semantic_grammars}). Whenever a new word or phrase is selected, the
system parses the sentence assembled so far, and processes the grammar to
determine the words/phrases that can be attached to the existing partial 
sentence to form a sentence the system can understand. These words/phrases
are then shown in the active menus. This way, only sentences the system
can understand can be input, and the user can browse through the
menus to explore the kinds of questions the system can handle.

Although a new semantic grammar is needed for each new knowledge
domain (see section \ref{portability}), {\sc Nlmenu} grammars do not have to
cope with a large number of possible user sentences. The menus can be
used to greatly restrict the kinds of sentences the user can input,
and hence the semantic grammars of {\sc Nlmenu} are relatively simple to
write.  According to \cite{Tennant2}, {\sc Nlmenu} is typically customised
for a new application in 1 -- 30 man-hours. 

The menu-based approach guarantees that the user's input is free from
spelling errors, and allows user queries to be input using a simple
pointing device. However, as mentioned in \cite{Tennant2}, in
applications requiring long questions menu-searching can become
tedious.

\section{NLIDBs as intelligent agents} \label{intelligent}

Most {\sc Nlidb}s described so far are direct interfaces to the
underlying database, in the sense that they simply translate user
questions to suitable database queries. The database is seen as the
only source of information about the world. (The {\sc Nlidb} itself
only contains a simplistic world-model -- section \ref{irls}).  This
section will discuss {\sc Nlidb}s that contain knowledge about the
surrounding world and about the user. This knowledge allows them to
handle questions that cannot be answered directly by the {\sc Dbms},
and to understand better the user's goals.

\subsection{Reasoning about the world} \label{domain_reasoning}

In some cases it may not be possible to answer a natural language
question, although all the necessary raw data are present in the database.
Questions involving common sense or domain expertise are typical
examples. In these cases, the answer to the question is not explicitly
stored in the database; to produce the answer, the {\sc Nlidb} must be able to
carry out reasoning based on the data stored in the database.

Let us assume, for example, that the database holds the medical
records af all patients in a hospital (e.g.\ the names of the
patients, their ages, diagnoses made for each patient, treatments the
patients have received, etc.), and let us assume that the {\sc Nlidb}
is asked \qit{Which patients are likely to develop lung disease?}. Let
us also assume that the database does not show explicitly how probable
it is for each patient to develop each disease (i.e.\ these
probabilities are not stored in the database). In this case the {\sc
  Nlidb} will not be able to answer the question, unless it is told
how to use the medical data in the database to determine how probable it
is for each patient to develop lung disease. In other words, the {\sc
  Nlidb} must have access to domain expertise (in this case medical
expertise, perhaps in the form of rules), showing how to carry out
reasoning based on the raw data in the database. 

\begin{figure}
\begin{center}
\mbox{\epsfysize=2in \epsffile{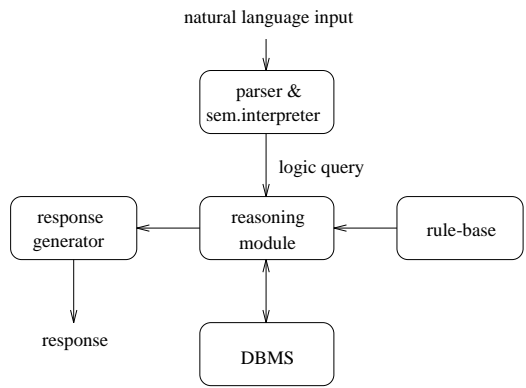}}
\caption{Architecture of a {\sc Nlidb} with a reasoning module}
\label{reasoning_arch}
\end{center}
\end{figure}

The architecture of figure \ref{irl_arch} can be extended to allow the
{\sc Nlidb} to carry out reasoning based on the data in the database.
One possible extension, exemplified by {\sc Loqui} \cite{Binot}, is to
add a rule-based reasoning module on top of the database as shown in
figure \ref{reasoning_arch}.

Systems adopting the architecture of figure \ref{reasoning_arch} still
transform the natural language question into a logic query that
captures the meaning of the question, as discussed in section
\ref{irls}.  When asked \qit{Which patients are likely to develop lung
  disease?}, such a system could generate the logic query (variables
start with capital letters):

\begin{verbatim}
report Patient such that:
   is_patient(Patient) and
   likely_to_develop(Patient, lung_disease)
\end{verbatim}

Some of the predicates in the logic queries may be linked directly to
tables in the database, as discussed in section \ref{irls}.  For
example, $is\_patient(Patient)$\/ could be linked to a database
table $patients$\/ listing all the patients in the hospital.
Thus, $is\_patient(Patient)$\/ would be true if
$patients$\/ showed that $Patient$\/ is a patient in the hospital.

For predicates not directly linked to database tables, suitable rules
must be present in the rule-base, showing how to evaluate instances of
these predicates. Continuing with the same example, the rule-base
could contain the rule:

\begin{verbatim}
likely_to_develop(Patient, Disease) if 
  prob(develop(Patient, Disease)) > 0.7
\end{verbatim}

\noindent The rule above says that a patient is likely to develop a
disease if the probability of the patient developing the disease is
greater than 0.7 

Other rules would show how to use the data in the database to assess
the probability of a patient developing a disease. For example:

\begin{verbatim}
if has(Patient, blood_disease) then
  prob(develop(Patient, lung_disease)) = 0.6
\end{verbatim}

\begin{verbatim}
if is_smoker(Patient) then
  prob(develop(Patient, lung_disease)) = 0.7
\end{verbatim}

\noindent
Finally, other rules would show how to combine the probabilities
predicted by the rules above, in cases where more than one rules apply
(e.g.\ patient with blood disease who is also a smoker). (See
\cite{Giarratano} for an introduction to rule-based reasoning with
probabilities.)

In {\sc Loqui}, according to \cite{Loqui}, both the logic queries and
the rules of the rule-base are expressed in Prolog. The logic query is
treated as a Prolog goal to be evaluated by the Prolog interpreter,
and rows in the database tables are treated as Prolog facts.  (See for
example \cite{Lucas} \cite{Ceri1} \cite{Ceri2} \cite{Draxler} for
information on how Prolog can be linked to external databases.) Thus,
the full power of Prolog is available when evaluating a query, or when
defining the rules in the rule-base, and the internal Prolog database
can be used to hold knowledge about the world that cannot be easily
encoded in the external relational database. {\sc Chat-80}
\cite{Warren} and {\sc Masque} \cite{Auxerre1} \cite{Auxerre2} offer
similar capabilities.

{\sc Intellect}, according to \cite{Intellect}, can be configured to
use the {\sc Kbms} \cite{KBMS} expert system shell as its reasoning
module. The logic query is passed to {\sc Kbms}, and it is the
responsibility of {\sc Kbms} to find all the possible answers using
its specialised inferencing techniques and the facts stored in the
external database. This way, the full power of the expert system shell
(e.g.\ forward chaining, backward chaining, hypothetical reasoning,
etc.) is available when evaluating the logic query, and the shell's
tools can be used to develop the rule-base.
The natural language capabilities of {\sc Intellect} can be used to
define {\sc Kbms} rules. For example \cite{KBMS}, a {\sc Kbms} rule
can be entered as \qit{{\sc IF} the customer has an overdue balance
  outstanding, {\sc Then} credit risk is high}. Natural language questions
can also query the shell's reasoning. For example, \qit{Why was the
  application of Smith rejected?}\/ would cause the system to report
the steps it took to reject Smith's application.

Perhaps the reasoning module should not be considered part of the {\sc
  Nlidb}, but rather an independent layer on top of the database
system (or a component of an ``expert database'' system). A strong
argument in favour of this view is that the ability to carry out
reasoning on top of the database is not needed only in {\sc Nlidb}s. A
graphical interface would also need to access the reasoning module in
order to answer a graphical query corresponding to \qit{Which patients
  are likely to develop lung disease?}. The reasoning module should
not, therefore, be part of the {\sc Nlidb}'s code. It should be an
independent module, available to any application that may need it.

\subsection{Reasoning about the user's goals} \label{user_model}

User requests often do not express literally what the user wants to know.
Let us consider the following dialogue, based on an example from
\cite{Johnson}:

\smallskip
\begin{tabular}{l}
\sys{>} \usr{Do American Airlines have a night flight to Dallas?}\\
\sys{yes}
\end{tabular}
\smallskip

\noindent It would be better if the system could respond:

\smallskip
\begin{tabular}{l}
\sys{Yes, flight 712.} \\ 
or: \\
\sys{No, but United have flight 655.}
\end{tabular}
\smallskip \\ In the case of the last two responses, instead of simply
printing a yes/no response, the {\sc Nlidb} has also reported additional
information that the user would probably be interested to know. Notice
that in the last response an alternative United flight was reported,
even though the user had requested an American Airlines flight.

As already mentioned in section \ref{response_generation}, in some
cases reasonably cooperative responses can be generated by using
relatively simple mechanisms. For example, in yes/no questions like
\qit{Do American Airlines have a night flight to Dallas?}, if the
answer is negative, the {\sc Nlidb} could report a ``\sys{no}'', along with
the answer to a ``weaker'' question like \qit{Does a carrier have a
  night flight to Dallas?}. The weaker question is generated from the
original one by removing one restriction (the restriction that the
carrier must be American Airlines in our example). (A similar but more
elaborate method is proposed in \cite{Kaplan2}.)

Unfortunately, simple strategies, like dropping an arbitrary
restriction, do not always generate acceptable responses. As pointed
out in \cite{Johnson}, the previous dialogue could have been:

\smallskip
\begin{tabular}{l}
\sys{>} \usr{Do American Airlines have a night flight to Dallas?}\\
\sys{No, but they have one to Miami.}
\end{tabular}
\smallskip \\
In the latter dialogue the system has dropped the restriction that the
flight must go to Dallas, rather than the restriction that the carrier
must be American Airlines. The resulting ``cooperative'' response
would probably be irritating for the user.
To avoid such inappropriate responses, the {\sc Nlidb} must be able to
``understand'' the user's goals. (This is often called ``plan
recognition'' in the literature -- see for example \cite{Allen}.) In
the previous question, we would like the {\sc Nlidb} to be able to understand
that the user's goal is to go to Dallas, and that therefore it would
be inappropriate to report a flight to Miami.  

This could be achieved by having a rule stating that when a person
asks about a flight to a particular location, the goal of the person
is probably to reach that location. Another rule could state that if
the person has specified a particular carrier, then flying with that
carrier is probably a preference but not a goal.  Finally, another
rule could state that if it is not possible to satisfy all goals and
preferences, then preferences, but not goals, can be dropped. Hence,
in \qit{Do American Airlines have a night flight to Dallas?}, if there
is no American Airlines night flight to Dallas, the system would drop
the carrier constraint, and not the destination constraint.

The reasoning about the user's goals could be carried out by reasoning
modules similar to the ones described in the previous section (figure
\ref{reasoning_arch}). In this case, however, the rule-base would
contain rules explaining how the user's requests relate to his/her
goals, and what to do to satisfy the user's goals.  The user's request
would be mapped to a logic query, and the logic query would be passed
to the reasoning module. The reasoning module would then try to satisfy
the user by reasoning about his/her goals, and by retrieving
information from the database.

The rules in the rule-base can be thought as constituting a {\em model
  of the user}, a logic system describing how user utterances relate
to the user's intentions and beliefs. The need for elaborate user
models becomes essential in dialogue systems, discussed in the
following section.

\subsection{Dialogue systems} \label{dialogue_systems}

The architectures discussed so far focus on the understanding of
individual user questions. The parsing and semantic interpretation modules
form the backbone of the system, while reasoning and response generation
modules are treated as peripheral accessories. In other words, it is the
processing of each individual question that drives the system's behaviour.

\begin{figure*}
\begin{center}
\mbox{\epsfxsize=5in \epsffile{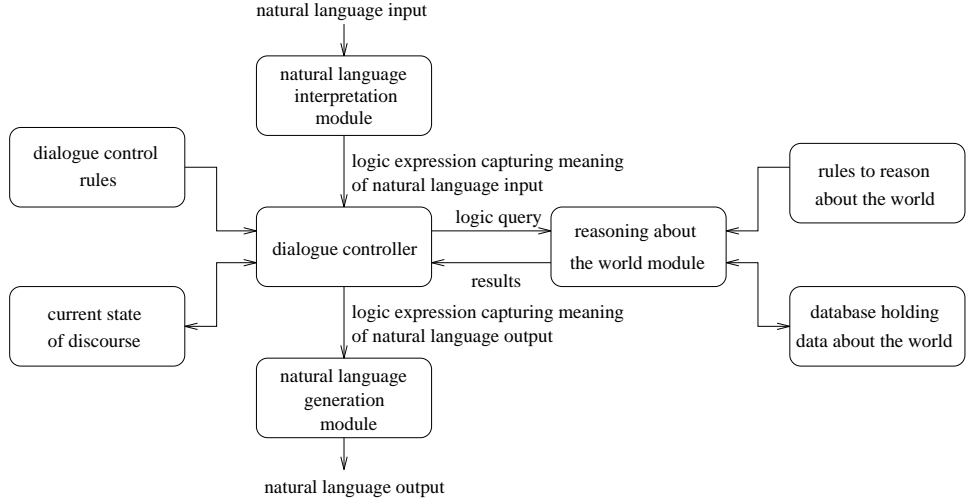}}
\caption{Possible architecture of a dialogue-oriented system}
\label{dialogue_arch}
\end{center}
\end{figure*}

In dialogue-oriented systems the system's behaviour is driven by a
reasoning module, which reasons about the goals and beliefs of both the
user and the system. A possible architecture of a dialogue-oriented
system is shown in figure \ref{dialogue_arch}. (The architecture shown is
greatly influenced by the {\sc Ir-nli }architecture described in \cite{Guida}.)

The central module of the architecture in figure \ref{dialogue_arch}
is the {\em dialogue controller}. The dialogue controller is a
reasoning module, whose purpose is to understand the user's needs and
to convey the relevant information to the user. The dialogue
controller consults a rule-base, that contains rules explaining how to
carry out a dialogue. 
For example, the following is a simplified dialogue-control rule for
{\sc Wisber} \cite{Gerlach}, a natural language consultation system:

{\small
\[
\begin{array}{lllll}
\begin{array}{l}
(want \; system \\
\; \; \; (know \; system \; X)) \; \wedge \\
(believe \; system \\
\; \; \; (know \; user \; (related \; X))) \; \wedge \\
\neg (know \; system \\
\; \; \; \neg (know \; user X))\\
\end{array}
& \! \! \! \! \!\! \Rightarrow \!\!\! \! \! \! &
\begin{array}{l}
     (ask \\
     system \\
     user \\
     X)
\end{array}
& \!\!\!\!\!\! \Rightarrow  \! \! \! \!\!\! & 
\begin{array}{l}
(know \; mutual \\
\; \; \; (want \; system \\
\; \; \; \; \; \; (know \; system \; X))) \; \wedge \\
(know \; mutual \\
\; \; \; (believe \; system \\
\; \; \; \; \; \; (know \; user \; (related \; X)))) \\
\end{array} 
\end{array}
\]
} 

\noindent The rule says that if:

\begin{itemize}

\item the system wants to know a piece of knowledge, called $X$, and

\item the system believes that the user knows something related to $X$,
and

\item the system does not know that the user does not know $X$

\end{itemize}
then, the system should ask the user about $X$, and after the question has
been asked, the system should assume that:
\begin{itemize}

\item both the system and the user know that the system wants to learn
  $X$, and

\item both the system and the user know that the system believes the
user knows something related to $X$.

\end{itemize}

In figure \ref{dialogue_arch} the {\em current state of discourse}\/
contains facts describing the current goals and beliefs of the system,
and facts describing what the system currently understands to be the
goals and beliefs of the user.

Both the {\em natural language interpretation}\/ and {\em natural
  language generation}\/ modules are subordinate to the dialogue
controller. Whenever the dialogue controller wishes to communicate
information to the user (e.g.\ ask the user a question, report retrieved
information), it formulates a logic expression that captures
the meaning of the message to be generated.  The logic expression is
then passed to the natural language generation module, which
transforms it into a natural language message. The natural language
interpretation module is then used to transform the user's response to
a logic expression, and this logic expression is passed to the
dialogue controller.

Finally, whenever the dialogue controller wishes to retrieve
information about the world, it formulates a logic
query which is passed to the {\em reasoning about the world}\/ module.
The latter uses the {\em rules to reason about the
world}\/ to carry out reasoning based on the data stored in the external
database (as discussed in section \ref{domain_reasoning}). 

In dialogue-oriented systems, the generation of the system's messages
(see section \ref{response_generation}) is equally important to the
interpretation of the user's input.  The system's messages must be
carefully chosen, to guide the dialogue towards areas that will
clarify the user's goals and beliefs. The system's messages must also
reflect correctly its understanding and its capabilities.

\section{Further Topics} \label{further_topics}

This section highlights some areas of {\sc Nlidb} research that
have been less explored, namely database updates, meta-knowledge
questions, temporal questions, and multi-modal interfaces.

\subsection{Database updates} \label{updates}

Although in several {\sc Nlidb}s the user is allowed to update the
interface's linguistic knowledge using natural language requests
(section \ref{knowledge_domain_portability}), few systems support
natural language requests to update the underlying database. Database
updates by natural language introduce several new problems, which will
be illustrated in this section.

In {\sc Ask} \cite{Thompson1} \cite{Thompson2} update requests can appear at
any point during the dialogue with the system. The following dialogue is based
on examples from \cite{Thompson1} \cite{Thompson2}:

\medskip
\noindent
\sys{>} \usr{What is the home port of the Tokyo Maru?} \\
\sys{Yokohama} \\
\sys{>} \usr{Home port of the Tokyo Maru is Hong Kong.} \\
\sys{Yokohama has been replaced by Hong Kong as home port of Tokyo Maru.} \\
\sys{>} \usr{The cargo of the Tokyo Maru is coal.} \\
\sys{Coal has been added as the cargo of}\\
\sys{Tokyo Maru.} \\
\sys{>} \usr{What is the cargo of the Tokyo Maru?} \\
\sys{Grain} \\
\sys{Coal}  
\smallskip

{\sc Ask} knows that each ship may only have one home port. Therefore,
when informed that \qit{Home port of the Tokyo Maru is Hong Kong.}, it
replaces \qit{Yokohama}\/ (previously known to be the home port of the
Tokyo Maru) by \qit{Hong Kong}. In contrast, a ship may carry more
than one kinds of cargo. Thus, when informed that \qit{The cargo of
  the Tokyo Maru is coal.}, {\sc Ask} adds \qit{Coal}\/ to the kinds
of cargo already known to be carried by the Tokyo Maru
(\qit{Grain}\/).
     
Unfortunately, satisfying a natural language update request is not
always that straightforward.  As explained in \cite{Davidson1}, 
natural language update requests often cannot be satisfied, or they are
ambiguous, or they lead to unanticipated side-effects, due to database
constraints the user is not aware of.
Consider a database containing the following tables (the example
is borrowed from \cite{Davidson1}):

\begin{center}
\begin{tabular}{lr}
\begin{tabular}[t]{llll}
\multicolumn{3}{c}{employees\_table} \\ 
employee & salary & department \\
\hline
Adams & 3000 & inventory \\
White & 3500 & marketing \\
Brown & 2500 & sales \\
Smith & 2500 & sales 
\end{tabular}
&
\begin{tabular}[t]{ll}
\multicolumn{2}{c}{departments\_table} \\
department & manager \\
\hline
sales & Jones \\
marketing & Baker \\
inventory & Fisher 
\end{tabular}
\end{tabular}
\end{center}

\noindent The following dialogue takes place:

\smallskip
\sys{>} \usr{List the employees and their managers.}\\
\sys{ \begin{tabular}{ll}
      employee & manager \\
      \hline
      Adams & Fisher \\
      White & Baker \\
      Brown & Jones \\
      Smith & Jones \\
      \end{tabular}} \\
\sys{>} \usr{Change Brown's manager from Jones to Baker.} \\
\sys{Done.} \\
\sys{>} \usr{List the employees and their managers.} \\
\sys{ \begin{tabular}{ll}
      employee & manager \\
      \hline
      Adams & Fisher \\
      White & Baker \\
      Brown & Baker \\
      Smith & Baker \\
      \end{tabular}
    } 
\medskip

Notice that Smith's manager has also changed from Jones to Baker, although 
this was not requested. To a user not aware of the database's structure, 
such a behaviour would appear erratic. The user does not know that in the
database employees are linked to managers through their departments. Since
Brown works in the sales department, the system has changed the sales 
manager from Jones to Baker, which has also caused other employees in the 
sales department (e.g.\ Smith) to get a new manager. 

Because of the indirect link between employees and managers, the
update request is actually ambiguous.  Instead of changing the sales
manager from Jones to Baker, the system could have moved Smith from
the sales department (managed by Jones) to the marketing department
(managed by Baker).

{\sc Piquet} \cite{Davidson1} maintains a connection graph, which
models the user's view of the database structure.  Whenever a user
utterance reflects awareness or presupposition of a database object,
link, restriction etc., the graph is modified accordingly.  Every time
an ambiguous update request is encountered, the system consults the
connection graph to determine which of the possible interpretations of
the update request are meaningful with respect to the user's current
view of the underlying database. It then ranks the remaining candidate
interpretations, using a set of heuristic rules. For example,
interpretations causing fewer side-effects with respect to the user's
view are preferred. Returning to the dialogue above, after \qit{List
  the employees and their managers.}\/ has been answered, the user's
view of the database is equivalent to a table of the form:

{\small
\begin{center}
\begin{tabular}{ll}
      employee & manager \\
      \hline
      Adams & Fisher \\
      White & Baker \\
      Brown & Jones \\
      Smith & Jones 
\end{tabular}
\end{center}
}

If the system interprets the request \qit{Change Brown's manager from
  Jones to Baker.}\/ as a request to change the sales manager from
Jones to Baker, then the user's view of the database becomes as shown
below on the left. (Side-effects are shown in {\bf this typeface}.)

{\small
\begin{center}
\begin{tabular}{ll}
\begin{tabular}{ll}
      employee & manager \\
      \hline
      Adams & Fisher \\
      White & Baker \\
      Brown & Baker \\
      Smith & {\bf Baker} 
\end{tabular}
&
\begin{tabular}{ll}
      employee & manager \\
      \hline
      Adams & Fisher \\
      White & Baker \\
      Brown & Baker \\
      Smith & Jones 
\end{tabular}
\end{tabular}
\end{center}
}

If the system interprets the request as a request to move Brown from
the sales department (managed by Jones) to the marketing department
(managed by Baker), then the user's view of the database becomes as
shown above on the right.  The second interpretation causes no
side-effects to the user's view, and thus it would have been preferred
by {\sc Piquet}'s heuristics.

\subsection{Meta-knowledge questions} \label{meta-questions}

Meta-knowledge questions are questions referring to knowledge about
knowledge. For example, some meta-knowledge questions could be
\cite{Hendrix1} \cite{Thompson2}:

\smallskip
\begin{tabular}{l}
\sys{>} \usr{What information is in the database?} \\
\sys{>} \usr{What is known about ships?} \\
\sys{>} \usr{What are the possible employee job titles?} \\
\sys{>} \usr{Can you handle relative clauses?} \\
\end{tabular}
\smallskip

The following example \cite{Thompson2} shows how {\sc Ask} reacts to one
meta-knowledge question:

\smallskip
\begin{tabular}{l}
\sys{>} \usr{What is known about ships?} \\
\sys{ \begin{tabular}{l}
      Some are in the following classes: navy, freighter, tanker\\
      All have the following attributes: destination, home port \\
      \dots
      \end{tabular}
    }
\end{tabular}
\smallskip

An interesting kind of meta-knowledge questions are {\em modal questions}. In
modal questions the user asks whether something can or must be the case.
For example:

\begin{center}
\begin{tabular}{l}
\sys{>} \usr{Can a female employee work in sales?} \\
\sys{>} \usr{Can an employee earn more than his manager?}\\
\sys{>} \usr{Is it true that all candidates must be over 20?}\\
\end{tabular}
\end{center}

\noindent In \cite{Lowden2} the authors present a method for handling modal
questions in {\sc Nlidb}s to relational databases. This method is used
in the {\sc Nlidb} developed at Essex University \cite{DeRoeck1}
\cite{DeRoeck2} \cite{Lowden1}.

\subsection{Temporal questions} \label{temporal_questions}

Most {\sc Nlidb}s were designed to interface to snapshot databases.
Snapshot databases only store information about one state of the
world, usually taken to be the ``present'' state. Consequently, most
{\sc Nlidb}s only support questions referring to the present state of
the world.
For example, a {\sc Nlidb} to a company's database would typically be able to
answer questions like:

\begin{center}
\begin{tabular}{l}
\sys{>} \usr{Who is the manager of the sales department?} \\
\sys{>} \usr{What is the maximum manager salary?}
\end{tabular}
\end{center}

\noindent but would usually not support temporal questions referring to
the past (or the future):

\begin{center}
\begin{tabular}{l}
\sys{>} \usr{Who was the previous manager of the sales department?} \\ 
\sys{>} \usr{What was the maximum manager salary during the last 10 years?} \\
\sys{>} \usr{While Smith was sales manager, did the annual sales
  income ever exceed}\\
\sys{ } \usr{1 million?}  
\end{tabular}
\end{center}

Most {\sc Nlidb}s cannot answer temporal questions because: (a) they
cannot cope with the semantics of natural language temporal
expressions (e.g.\ tenses/aspects, temporal subordinators), and (b)
they were designed to interface to "snapshot" databases, that do not
facilitate the manipulation of time-dependent information.

In the last decade, computer scientists have become increasingly
interested in {\em temporal database systems}, i.e.\ database systems
designed to store information about previous or future states of the
world, and to generally support the notion of time.  (The reader is
referred to \cite{Tansel3} for an introduction to temporal databases.)
Clifford \cite{Clifford} \cite{Clifford2} \cite{Clifford3}, Hafner
\cite{Hafner3}, and Hinrichs \cite{Hinrichs1} are among those who have
considered natural language interfaces to temporal
databases.\footnote{The authors are also currently involved in
  research on natural language interfaces to temporal databases.}

\subsection{Multimodal interfaces} \label{multimodal}

As discussed in section \ref{comparison}, form-based interfaces and
graphical interfaces appear to have (at least some) advantages over
natural language interfaces. Some systems attempt to merge natural
language with graphics, menus, and forms, to combine the strengths of
all modalities.

In {\sc Janus} \cite{Bobrow2} the user is allowed to combine text, menus,
graphics, and speech when forming his/her queries. A user could, for
example, type the question \qit{Which of those submarines has the
  greatest probability of locating A within 10 hours?}, and then use
the mouse and a screen displaying the positions of ships to determine
the set of ships to which \qit{those}\/ refers.

One way to integrate natural language and other modalities is
exemplified by {\sc Shoptalk} \cite{Cohen}. {\sc Shoptalk} is actually
a complex manufacturing information and decision support system with
multimodal input capabilities, rather than a simple {\sc Nlidb}. The
methods, however, used in {\sc Shoptalk} could also be used in simple
{\sc Nlidb}s.  The following examples are all borrowed from
\cite{Cohen}.

{\sc Shoptalk} initially displays a set of menus, listing the available kinds
of actions and queries (e.g. \sys{move a lot}, \sys{put an object} etc).
This way, the user has a clear view of the system's capabilities. 
When an action/query-type has been selected, {\sc Shoptalk} displays a form
containing fields that have to be filled in. For example, if the user
selects the \sys{move\_lots} action, the system displays the form:

\begin{center}
\begin{tabular}{ll}
\multicolumn{2}{c}{\sys{move\_lots}} \\
\hline
\sys{which lots:} & \\
\sys{to which station:} & \\
\sys{whenever:} & 
\end{tabular}
\end{center}

Each field can be filled using input in the form of any modality
supported by {\sc Shoptalk}. For example, the \sys{which lots} field could be
filled by selecting lots from a diagram displayed on screen, or by
typing natural language. The form above, could be filled as follows
(\usr{$<$here$>$} denotes pointing with the mouse):

\begin{center}
\begin{tabular}{ll}
\multicolumn{2}{c}{\sys{move\_lots}} \\
\hline
\sys{which lots:}&\usr{each baked lot} \\
\sys{to which station:}&\usr{the manual inspection station}\\
&\usr{with the smallest queue} \\
\sys{whenever:}&\usr{it has been moved $<$here$>$} 
\end{tabular}
\end{center}

By allowing both natural language and mouse-pointing, the user can
combine the strengths of both modalities. In the case of the
\sys{which lots} field, the user has chosen natural language
(\qit{each baked lot}\/), which allows universal quantification
(\qit{each}\/) to be expressed easily. It would be difficult to
express universal quantification using graphics and the mouse: the
system would have to display all the baked lots, and the user would
have to select all the baked lots from the display using the mouse. 

Similarly, natural language allows \qit{the manual inspection station
  with the smallest queue}\/ to be expressed easily and concisely.
Without natural language support, the user would probably have to use
a formal specification language, or to use numerous nested menus.
In contrast, in the case of the \sys{whenever} field, the user has
substituted part of the natural language expression by a pointing action
on a diagram, thus avoiding the need to provide a lengthy natural language
description of the location.

The form-filling approach used in {\sc Shoptalk} also by-passes some kinds
of natural language ambiguity.  In the following simplified example
from \cite{Cohen},

\begin{center}
\begin{tabular}{ll}
\multicolumn{2}{c}{\sys{put}} \\
\hline
\sys{object:} & \usr{the block in the box} \\
\sys{destination:} & \usr{on the table in the corner}
\end{tabular}
\end{center}

\noindent it is clear to the system that \qit{in the box}\/ is a modifier
of \qit{the block}, that \qit{in the corner}\/ is a modifier of
\qit{the table}, and that \qit{on the table in the corner}\/ is the
description of the destination and not, for example, a modifier of the
object to be moved (see section \ref{modif_att} for a description of
the modifier attachment problem).

After each question, {\sc Shoptalk} graphically displays objects to which
anaphoric expressions (e.g.\ pronouns -- see section \ref{anaphora})
of follow-up questions may refer. The user can subsequently deactivate
some of these possible referents by pointing on their graphical
representations, thus reducing the number of possible referents the
system has to consider in follow-up questions.

\begin{figure}
\begin{center}
\mbox{\epsfysize=.8in\epsffile{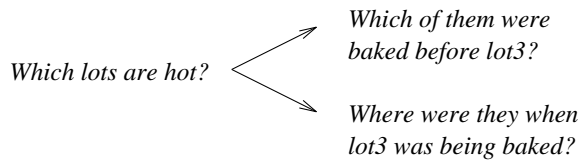}}
\caption{A context graph in {\sc Shoptalk}}
\label{shop_context}
\end{center}
\end{figure}

{\sc Shoptalk} also displays a graphical representation of the discourse
structure, allowing the user to specify the context to which follow up
questions refer. Figure \ref{shop_context} shows a modified example from
\cite{Cohen}. The user has specified that both \qit{Which of them were baked
before lot3?}\/ and \qit{Where were they when lot3 was being baked?}\/ are
follow-up questions to \qit{Which lots are hot?}. Therefore, the pronoun
\qit{they} in \qit{Where were they when lot3 was being baked?}\/ refers to
the hot lots, not to the hot lots that were baked before lot3.

Finally, {\sc Shoptalk} supports questions referring to past events or
states.  The user is allowed to specify the previous time to which a
query refers by either dragging a ``time-slider'', or by filling
special \sys{when} fields (see \cite{Cohen} for more details).

\section{The state of the art} \label{state_of_the_art}

This paper has attempted to serve two purposes: to introduce the
reader to the field of {\sc Nlidb}s by describing some of the central
issues, and to indicate the current situation in this area by
outlining the facilities and methods of typical implemented systems.
The goal of surveying the field can be achieved only incompletely at
any given moment. Nevertheless, our summary here is sufficient to give
rise to some general observations about recent work on {\sc Nlidb}s. 

We have not dealt in detail  with the evaluation of {\sc Nlidb}s. 
This is a complex problem. It is difficult enough to formulate general
quantitative (or subtle qualitative) measures of the effectiveness
of parsers (i.e.\ programs which extract syntactic information from
sentences); when we have to consider an entire system, involving 
semantics, querying, response generation, and user interfacing, the
situation is much more complicated. Hence, no standard benchmarks
have yet been developed (see \cite{Palmer90} for some discussion).
Under these circumstances, any appraisal of the current state of the
field must be impressionistic and subjective.

In order to structure our presentation round issues, we have
deliberately overlooked one dimension of distinction which might be
held to be important, namely, the extent to which the systems are
experimental research vehicles or working commercial applications.
The majority of the systems we have referred to are oriented more
towards exploration of possible techniques than towards the production
of marketable products. Those which seem to fall into the category of
experimental prototypes include {\sc Lunar}, {\sc Chat-80}, {\sc
  Masque}, {\sc Team}, {\sc Ask}, {\sc Eufid}, {\sc Ldc}, {\sc Tqa},
and others, while {\sc Intellect}, {\sc Parlance}, {\sc
  Languageaccess}, Q\&A, {\sc Natural Language}, {\sc Loqui}, and {\sc
  English Wizard} are actual products or systems used in real
applications.

In the context of these caveats, the following observations seem to us
to be valid:

\begin{enumerate}

\item There are several facets of {\sc Nlidb}s (e.g.\ dictionaries,
  parsing) where there has been a great deal of research and
  development, and where practical system modules can be constructed
  fairly straightforwardly.
\item Even in the relatively well-understood subareas mentioned above,
  there is not a single agreed consensus theory or technique.  In the
  other aspects of {\sc Nlidb}s, there are several difficult
  theoretical problems which remain to be solved before {\sc Nlidb}s
  will be fully useful and easily constructed.
\item The development of {\sc Nlidb}s is no longer as fashionable a
  topic within academic research as it was in the 1980s.
\item Although there are several commercially available systems which
  purport to allow querying of databases in English, it must be
  remembered that this claim is flexible in its possible
  interpretations.\footnote{As recently as 1986, a major manufacturer
    advertised a system which allowed you to program it ``in English'';
    on closer inspection this turned out to be an interpreter for the
    programming language Basic!} These systems may be extremely
  effective tools for database access, and may use some form of
  English (or other natural languages).  This does not mean that the
  problem of completely unrestricted use of natural language queries
  has been solved.
\item Interfaces which solely use keyboard input of natural language
  are not likely to be hugely practical; in the long term, the use of
  spoken input, and/or integration with graphical user interfaces, are
  more likely to be the route to practical success.  However,
  keyboard-based {\sc Nlidb}s are still a useful testbed for developing the
  necessary techniques (e.g.\ semantic interpretation) for subsequent
  incorporation into these more complex systems.
\end{enumerate}

\section*{Acknowledgments}

{\small 
The authors would like to thank Lieve Debille (BIM), Kathleen Pierco
(BIM), Madeleine Bates (BBN Systems and Technologies), Barclay
Rockwood (Natural Language Inc.), John Menkart (Trinzic), Hilary Quick
(IBM UK), and Bruce Bovill (SAS Software Ltd.) for providing
information about systems developed by their companies. The authors
are also grateful to Chris McNeilly, David Boltz, and John Nerbonne,
who provided pointers to the literature and addresses of companies
that have developed {\sc Nlidb}s. Special thanks to John Nerbonne for
comments on an earlier draft of this paper.

This paper reports work carried out by the first author under the
supervision of the other two authors. The first author is grateful to
the Greek State Scholarships Foundation for funding his studies in
Edinburgh.
}

\bibliographystyle{plain}
\bibliography{biblio}

\begin{thebibliography}{100}

\bibitem{Allen}
J.~Allen.
\newblock {Recognizing Intentions from Natural Language Utterances}.
\newblock In M.~Brady and R.C. Berwick, editors, {\em {Computational Models of
  Discourse}}, chapter~2, pages 107--166. MIT Press, Cambridge, Massachusetts,
  1983.

\bibitem{Alshawi}
H.~Alshawi.
\newblock {\em The {Core} {L}anguage {E}ngine}.
\newblock MIT Press, Cambridge, Massachusetts, 1992.

\bibitem{Alshawi2}
H.~Alshawi, D.~Carter, R.~Crouch, S.~Pulman, M.~Rayner, and A.~Smith.
\newblock {CLARE -- A Contextual Reasoning and Cooperative Response Framework
  for the Core Language Engine}.
\newblock Final report, SRI International, December 1992.

\bibitem{Androutsopoulos}
I.~Androutsopoulos.
\newblock {Interfacing a Natural Language Front-End to a Relational Database
  (MSc thesis)}.
\newblock Technical paper~11, Department of Artificial Intelligence, University
  of Edinburgh, 1993.

\bibitem{Androutsopoulos3}
I.~Androutsopoulos, G.~Ritchie, and P.~Thanisch.
\newblock {An Efficient and Portable Natural Language Query Interface for
  Relational Databases}.
\newblock In P.W. Chung, G.~Lovegrove, and M.~Ali, editors, {\em {Proceedings
  of the 6th International Conference on Industrial \& Engineering Applications
  of Artificial Intelligence and Expert Systems, Edinburgh, U.K.}}, pages
  327--330. Gordon and Breach Publishers Inc., Langhorne, PA, U.S.A., June
  1993.
\newblock ISBN 2--88124--604--4.

\bibitem{Auxerre1}
P.~Auxerre.
\newblock {MASQUE} {M}odular {A}nswering {S}ystem for {Q}ueries in {E}nglish -
  {P}rogrammer's {M}anual.
\newblock Technical Report AIAI/SR/11, Artificial Intelligence Applications
  Institute, University of Edinburgh, March 1986.

\bibitem{Auxerre2}
P.~Auxerre and R.~Inder.
\newblock {MASQUE} {M}odular {A}nswering {S}ystem for {Q}ueries in {E}nglish -
  {U}ser's {M}anual.
\newblock Technical Report AIAI/SR/10, Artificial Intelligence Applications
  Institute, University of Edinburgh, June 1986.

\bibitem{Ballard3}
B.~Ballard and D.~Stumberger.
\newblock {Semantic Acquisition in TELI}.
\newblock In {\em Proceedings of the 24th Annual Meeting of ACL, New York},
  pages 20--29, 1986.

\bibitem{Ballard1}
B.W. Ballard.
\newblock The {S}yntax and {S}emantics of {U}ser-{D}efined {M}odifiers in a
  {T}ransportable {N}atural {L}anguage {P}rocessor.
\newblock In {\em Proceedings of the 22nd Annual Meeting of ACL, Stanford,
  California}, pages 52--56, 1984.

\bibitem{Ballard2}
B.W. Ballard, J.C. Lusth, and N.L. Tinkham.
\newblock {LDC-1}: {A} {T}ransportable, {K}nowledge-based {N}atural {L}anguage
  {P}rocessor for {O}ffice {E}nvironments.
\newblock {\em ACM Transactions on Office Information Systems}, 2(1):1--25,
  January 1984.

\bibitem{Bates2}
M.~Bates, M.G. Moser, and D.~Stallard.
\newblock {The IRUS transportable natural language database interface}.
\newblock In L.~Kerschberg, editor, {\em {Expert Database Systems}}, pages
  617--630. Benjamin/Cummings, Menlo Park, CA., 1986.

\bibitem{BBN}
BBN Systems and Technologies.
\newblock {\em {BBN} {P}arlance {I}nterface {S}oftware -- {S}ystem {O}verview},
  1989.

\bibitem{Bell}
J.E. Bell and L.A. Rowe.
\newblock An {E}xploratory {S}tudy of {A}d {H}oc {Q}uery {L}anguages to
  {D}atabases.
\newblock In {\em Proceedings of the 8th {I}nternational {C}onference on {D}ata
  {E}ngineering, Tempe, Arizona}, pages 606--613. IEEE Computer Society Press,
  February 1992.

\bibitem{Loqui}
{BIM Information Technology}.
\newblock {\em {Loqui: An Open Natural Query System -- General Description}},
  1991.
\newblock (Commercial leaflet).

\bibitem{Binot}
J.-L. Binot, L.~Debille, D.~Sedlock, and B.~Vandecapelle.
\newblock Natural {L}anguage {I}nterfaces: {A} {N}ew {P}hilosophy.
\newblock {\em SunExpert Magazine}, pages 67--73, January 1991.

\bibitem{Bobrow1}
R.J. Bobrow.
\newblock The {RUS} {S}ystem.
\newblock In {\em Research in {N}atural {L}anguage {U}nderstanding, BBN Report
  3878}. Bolt Beranek and Newman Inc., Cambridge, Massachusetts, 1978.

\bibitem{Bobrow2}
R.J. Bobrow, P.~Resnik, and R.M. Weischedel.
\newblock {Multiple Underlying Systems: Translating User Requests into Programs
  to Produce Answers}.
\newblock In {\em Proceedings of the 28th Annual Meeting of ACL, Pittsburgh,
  Pennsylvania}, pages 227--234, 1990.

\bibitem{Capindale1990}
R.A. Capindale and R.G. Crawford.
\newblock {Using a Natural Language Interface with Casual Users}.
\newblock {\em International Journal of Man-Machine Studies}, 32:341--361,
  1990.

\bibitem{Carbonell}
J.G. Carbonell.
\newblock Discourse {P}ragmatics and {E}llipsis {R}esolution in
  {T}ask-{O}riented {N}atural {L}anguage {I}nterfaces.
\newblock In {\em Proceedings of the 21st Annual Meeting of ACL, Cambridge,
  Massachusetts}, pages 164--168, 1983.

\bibitem{Ceri2}
S.~Ceri, G.~Gottlob, and L.~Tanca.
\newblock {\em Logic {P}rogramming and {D}atabases}.
\newblock Springer-Verlag, Berlin, 1990.

\bibitem{Ceri1}
S.~Ceri, G.~Gottlob, and G.~Wiederhold.
\newblock Efficient {D}atabase {A}ccess from {P}rolog.
\newblock {\em {IEEE} {T}ransactions on {S}oftware {E}ngineering},
  15(2):153--163, February 1989.

\bibitem{Ceri3}
S.~Ceri and G.~Pelagatti.
\newblock {\em {Distributed Databases: Principles and Systems}}.
\newblock McGraw-Hill, New York, 1984.

\bibitem{Clifford3}
J.~Clifford.
\newblock {Natural Language Querying of Historical Databases}.
\newblock {\em Computational Linguistics}, 14(4):10--34, December 1988.

\bibitem{Clifford}
J.~Clifford.
\newblock {\em {Formal Semantics and Pragmatics for Natural Language
  Querying}}.
\newblock Cambridge Tracts in Theoretical Computer Science, Cambridge
  University Press, Cambridge, England, 1990.

\bibitem{Clifford2}
J.~Clifford and D.S. Warren.
\newblock {Formal Semantics for Time in Databases}.
\newblock {\em ACM Transactions on Database Systems}, 8(2):215--254, June 1983.

\bibitem{Codd1970}
E.F. Codd.
\newblock {A Relational Model for Large Shared Data Banks.}
\newblock {\em Communications of the ACM}, 13(6):377--387, 1970.

\bibitem{Codd1}
E.F. Codd.
\newblock {Seven Steps to RENDEZVOUS with the Casual User}.
\newblock In J.~Kimbie and K.~Koffeman, editors, {\em {Data Base Management}}.
  North-Holland Publishers, 1974.

\bibitem{Cohen}
P.R. Cohen.
\newblock The {R}ole of {N}atural {L}anguage in a {M}ultimodal {I}nterface.
\newblock Technical Note 514, Computer Dialogue Laboratory, SRI International,
  1991.

\bibitem{Copestake}
A.~Copestake and K.~Sparck~Jones.
\newblock {Natural Language Interfaces to Databases}.
\newblock {\em The Knowledge Engineering Review}, 5(4):225--249, 1990.

\bibitem{Damerau2}
F.~Damerau.
\newblock Operating statistics for the transformational question answering
  system.
\newblock {\em American Journal of Computational Linguistics}, 7:30--42, 1981.

\bibitem{Damerau}
F.~Damerau.
\newblock Problems and {S}ome {S}olutions in {C}ustomization of {N}atural
  {L}anguage {F}ront {E}nds.
\newblock {\em ACM Transactions on Office Information Systems}, 3(2):165--184,
  April 1985.

\bibitem{Davidson1}
J.~Davidson and S.J. Kaplan.
\newblock Natural {L}anguage {A}ccess to {D}ata {B}ases: {I}nterpreting
  {U}pdate {R}equests.
\newblock {\em Computational Linguistics}, 9(2):57--68, April--June 1983.

\bibitem{DeRoeck2}
A.N. De~Roeck, C.J. Fox, B.G.T. Lowden, R.~Turner, and B.R. Walls.
\newblock A {N}atural {L}anguage {S}ystem {B}ased on {F}ormal {S}emantics.
\newblock In {\em {Proceedings of the International Conference on Current
  Issues in Computational Linguistics, Pengang, Malaysia}}, 1991.

\bibitem{DeRoeck1}
A.N. De~Roeck, H.E. Jowsey, B.G.T. Lowden, R.~Turner, and B.R. Walls.
\newblock A {N}atural {L}anguage {F}ront {E}nd to {R}elational {S}ystems
  {B}ased on {F}ormal {S}emantics.
\newblock In {\em Proceedings of {I}nfoJapan '90, {T}okyo, {J}apan}, 1990.

\bibitem{DeRoeck3}
A.N. De~Roeck and B.G.T. Lowden.
\newblock {Generating English Paraphrases from Formal Relational Calculus
  Expressions}.
\newblock In {\em {Proceedings of the 11th International Conference on
  Computational Linguistics, Bonn, Germany}}, August 1986.

\bibitem{Dekleva1994}
S.M. Dekleva.
\newblock {Is Natural Language Querying Practical?}
\newblock {\em Data Base}, pages 24--36, May 1994.

\bibitem{Diaper1986}
D.~Diaper.
\newblock {Identifying the Knowledge Requirements of an Expert System's Natural
  Language Processing Interface}.
\newblock In M.D. Harrison and A.F. Monk, editors, {\em {People and Computers:
  Designing for Usability -- Proceedings of the Second Conference of the
  British Computer Society, Human Computer Interaction Specialist Group,
  University of York}}, pages 263--280. Cambridge University Press, September
  1986.

\bibitem{Dowty}
D.R. Dowty, R.E. Wall, and S.~Peters.
\newblock {\em {Introduction to Montague Semantics}}.
\newblock D.Reidel Publishing Company, Dordrecht, Holland, 1981.

\bibitem{Draxler}
C.~Draxler.
\newblock {\em {A}ccessing {R}elational and {H}igher {D}atabases {T}hrough
  {D}atabase {S}et {P}redicates in {Logic} {P}rogramming {L}anguages}.
\newblock PhD thesis, University of Zurich, 1992.

\bibitem{Epstein}
S.S. Epstein.
\newblock Transportable {N}atural {L}anguage {P}rocessing {T}hrough
  {S}implicity -- the {PRE} {S}ystem.
\newblock {\em ACM Transactions on Office Information Systems}, 3(2):107--120,
  April 1985.

\bibitem{Gal1991}
A.~Gal, G.~Lapalme, P.~Saint-Dizier, and H.~Somers.
\newblock {\em {Prolog for Natural Language Processing}}.
\newblock Wiley, Chichester, England., 1991.

\bibitem{Gazdar2}
G.~Gazdar and C.~Mellish.
\newblock {\em {Natural Language Processing in Lisp}}.
\newblock Addison-Wesley, 1989.
\newblock Prolog version also available.

\bibitem{Gazdar1}
G.~Gazdar, G.~Pullum, and I.~Sag.
\newblock {\em Generalized Phrase Structure Grammar}.
\newblock Blackwell, Oxford, 1985.

\bibitem{Gerlach}
M.~Gerlach and H.~Horacek.
\newblock Dialog {C}ontrol in a {N}atural {L}anguage {S}ystem.
\newblock In {\em Proceedings of the 4th Conference of the European Chapter of
  ACL, Manchester, England}, pages 27--34, 1989.

\bibitem{Giarratano}
J.G. Giarratano and G.~Riley.
\newblock {\em {Expert Systems: Principles and Programming}}.
\newblock PWS--Kent Publishing Company, Boston, 1989.
\newblock ISBN 0-87835-335-6.

\bibitem{Ginsparg}
J.M. Ginsparg.
\newblock A {R}obust {P}ortable {N}atural {L}anguage {D}atabase {I}nterface.
\newblock In {\em Proceedings of the 1st Conference on Applied Natural Language
  Processing, Santa Monica, California}, pages 25--30, 1983.

\bibitem{Grosz1}
B.J. Grosz.
\newblock {TEAM}: {A} {T}ransportable {N}atural-{L}anguage {I}nterface
  {S}ystem.
\newblock In {\em Proceedings of the 1st Conference on Applied Natural Language
  Processing, Santa Monica, California}, pages 39--45, 1983.

\bibitem{Grosz2}
B.J. Grosz, D.E. Appelt, P.A. Martin, and F.C.N. Pereira.
\newblock {TEAM}: {A}n {E}xperiment in the {D}esign of {T}ransportable
  {N}atural-{L}anguage {I}nterfaces.
\newblock {\em Artificial Intelligence}, 32:173--243, 1987.

\bibitem{Guida}
G.~Guida and C.~Tasso.
\newblock {IR-NLI}: {A}n {E}xpert {N}atural {L}anguage {I}nterface to {O}nline
  {D}ata {B}ases.
\newblock In {\em Proceedings of the 1st Conference on Applied Natural Language
  Processing, Santa Monica, California}, pages 31--38, 1983.

\bibitem{Hafner2}
C.D. Hafner.
\newblock Interaction of {K}nowledge {S}ources in a {P}ortable {N}atural
  {L}anguage {I}nterface.
\newblock In {\em Proceedings of the 22nd Annual Meeting of ACL, Stanford,
  California}, pages 57--60, 1984.

\bibitem{Hafner3}
C.D. Hafner.
\newblock {Semantics of Temporal Queries and Temporal Data}.
\newblock In {\em Proceedings of the 23rd Annual Meeting of ACL, Chicago,
  Illinois}, pages 1--8, July 1985.

\bibitem{Hafner1}
C.D Hafner and K.~Godden.
\newblock Portability of {S}yntax and {S}emantics in {D}atalog.
\newblock {\em ACM Transactions on Office Information Systems}, 3(2):141--164,
  April 1985.

\bibitem{Harris3}
L.R. Harris.
\newblock {User-oriented Data Base Query with the ROBOT Natural Language Query
  System}.
\newblock {\em International Journal of Man-Machine Studies}, 9:697--713, 1977.

\bibitem{Harris2}
L.R. Harris.
\newblock {The ROBOT System: Natural Language Processing Applied to Data Base
  Query}.
\newblock In {\em {Proceedings of the ACM'78 Annual Conference}}, 1978.

\bibitem{Harris1}
L.R. Harris.
\newblock {Experience with ROBOT in 12 Commercial Natural Language Data Base
  Query Applications}.
\newblock In {\em {Proceedings of the 6th International Joint Conference on
  Artificial Intelligence, Tokyo, Japan}}, pages 365--368, 1979.

\bibitem{Harris4}
L.R. Harris.
\newblock {Experience with INTELLECT: Artificial Intelligence Technology
  Transfer}.
\newblock {\em The AI Magazine}, 5(2):43--50, 1984.

\bibitem{Hendrix1}
G.~Hendrix.
\newblock Natural {L}anguage {I}nterface (panel).
\newblock {\em Computational Linguistics}, 8(2):55--61, April--June 1982.

\bibitem{Hendrix2}
G.~Hendrix, E.~Sacerdoti, D.~Sagalowicz, and J.~Slocum.
\newblock {Developing a Natural Language Interface to Complex Data}.
\newblock {\em ACM Transactions on Database Systems}, 3(2):105--147, 1978.

\bibitem{Hinrichs1}
E.W. Hinrichs.
\newblock Tense, {Q}uantifiers, and {C}ontexts.
\newblock {\em Computational Linguistics}, 14(2):3--14, June 1988.

\bibitem{Hirst}
G.~Hirst.
\newblock {\em {Anaphora in Natural Language Understanding: A Survey}}.
\newblock Springer-Verlag, Berlin, 1981.

\bibitem{Hobbs1986}
J.R. Hobbs.
\newblock {Resolving Pronoun References}.
\newblock In B.J. Grosz, K.~Sparck~Jones, and B.L. Webber, editors, {\em
  {Readings in Natural Language Processing}}, pages 339--352. Morgan Kaufmann
  Publishers, California, 1986.

\bibitem{Jarke}
M.~Jarke, J.A. Turner, E.A. Stohr, Y.~Vassiliou, N.H. White, and K.~Michielsen.
\newblock {A Field Evaluation of Natural Language for Data Retrieval}.
\newblock {\em IEEE Transactions on Software Engineering}, SE-11(1):97--113,
  1985.

\bibitem{Johnson}
T.~Johnson.
\newblock {\em Natural {L}anguage {C}omputing: {T}he {C}ommercial
  {A}pplications}.
\newblock Ovum Ltd., London, 1985.

\bibitem{Kaplan}
S.J. Kaplan.
\newblock {Cooperative Responses from a Portable Natural Language Data Base
  Query System}.
\newblock {\em Artificial Intelligence}, 19:165--187, 1982.

\bibitem{Kaplan2}
S.J. Kaplan.
\newblock {Cooperative Responses from a Portable Natural Language Database
  Query System}.
\newblock In M.~Brady and R.C. Berwick, editors, {\em {Computational Models of
  Discourse}}, chapter~3, pages 167--208. MIT Press, Cambridge, Massachusetts,
  1983.

\bibitem{Lopes}
G.P. Lopes.
\newblock Transforming {E}nglish {I}nterfaces to {O}ther {L}anguages: {A}n
  {E}xperiment with {P}ortoguese.
\newblock In {\em Proceedings of the 22nd Annual Meeting of ACL, Stanford,
  California}, 1984.

\bibitem{Lowden3}
B.G.T. Lowden and A.N. De~Roeck.
\newblock {REMIT: A Natural Language Paraphraser for Relational Query
  Expressions}.
\newblock {\em ICL Technical Journal}, 5(1):32--45, May 1986.

\bibitem{Lowden2}
B.G.T. Lowden, B.R. Walls, A.N. De~Roeck, Fox C.J., and R.~Turner.
\newblock Modal {R}easoning in {R}elational {S}ystems.
\newblock Technical Report CSM-163, University of Essex, Dept. of Computer
  Science, 1991.

\bibitem{Lowden1}
B.G.T. Lowden, B.R. Walls, A.N. De~Roeck, C.J. Fox, and R.~Turner.
\newblock {A Formal Approach to Translating English into SQL}.
\newblock In Jackson and Robinson, editors, {\em Proceedings of the 9th British
  National Conference on Databases (BNCOD)}, 1991.

\bibitem{Lucas}
R.~Lucas.
\newblock {\em {D}atabase {A}pplications {U}sing {P}rolog}.
\newblock Halsted Press, 1988.

\bibitem{Main}
M.G. Main and D.B. Benson.
\newblock Denotational {S}emantics for ``{N}atural {L}anguage'' {Q}uestion
  {A}nswering {P}rograms.
\newblock {\em Computational Linguistics}, 9(1):11--21, January--March 1983.

\bibitem{Manferdelli}
J.L. Manferdelli.
\newblock {Natural Languages}.
\newblock {\em Sun Technology}, pages 122--129, Summer 1989.

\bibitem{Martin}
P.~Martin, D.~Appelt, and F.~Pereira.
\newblock {Transportability and Generality in a Natural-Language Interface
  System}.
\newblock In B.J. Grosz, K.~Sparck~Jones, and B.L. Webber, editors, {\em
  {Readings in Natural Language Processing}}, pages 585--593. Morgan Kaufmann
  Publishers, California, 1986.

\bibitem{Martin1981}
W.A. Martin, K.W. Church, and R.S. Patil.
\newblock {Preliminary Analysis of a Breadth-First Parsing Algorithm:
  Theoretical and Experimental Results}.
\newblock Technical report MIT/LCS/TR-261, Laboratory for Computer Science,
  Massachusetts Institute of Technology, June 1981.

\bibitem{Natural_Language}
Natural Language Inc.
\newblock {\em {Natural Language 5.0}}.
\newblock (Commercial leaflets).

\bibitem{Ott}
N.~Ott.
\newblock Aspects of the {A}utomatic {G}eneration of {SQL} {S}tatements in a
  {N}atural {L}anguage {Q}uery {I}nterface.
\newblock {\em Information Systems}, 17(2):147--159, 1992.

\bibitem{Ovum1991}
Ovum Ltd., London.
\newblock {\em {Natural Language Markets}}, 1991.
\newblock ISBN 0-903 969 610.

\bibitem{Palmer90}
M.~Palmer and T.~Finin.
\newblock {Workshop on the Evaluation of Natural Language Processing Systems}.
\newblock {\em Computational Linguistics}, 16(3):175--181, 1990.

\bibitem{Perrault}
C.R. Perrault and B.J. Grosz.
\newblock Natural {L}anguage {I}nterfaces.
\newblock In H.E. Shrobe, editor, {\em Exploring {A}rtificial {I}ntelligence},
  pages 133--172. Morgan Kaufmann Publishers Inc., San Mateo, California, 1988.

\bibitem{Ingres1}
Relational Technology Inc.
\newblock {\em {Using INGRES through Forms and Menus}}, 1989.

\bibitem{Resnik}
P.~Resnik.
\newblock {Access to Multiple Underlying Systems in JANUS}.
\newblock BBN report 7142, Bolt Beranek and Newman Inc., Cambridge,
  Massachusetts, September 1989.

\bibitem{Scha}
R.J.H. Scha.
\newblock {Philips Question Answering System PHILIQA1}.
\newblock In {\em SIGART Newsletter, no.61}. ACM, New York, February 1977.

\bibitem{Sentance}
S.~Sentance.
\newblock Improved {R}esponses from an {E}nglish {L}anguage {F}ront {E}nd.
\newblock Master's thesis, Dept. of Artificial Intelligence, University of
  Edinburgh, 1989.

\bibitem{Shieber}
S.M. Shieber.
\newblock {\em {An Introduction to Unification-Based Approaches to Grammar}}.
\newblock CSLI Lecture Notes Number 4, Centre for the Study of Language and
  Information, Stanford, California, 1986.

\bibitem{Sijtsma}
W.~Sijtsma and O.~Zweekhorst.
\newblock {Comparison and Review of Commercial Natural Language Interfaces}.
\newblock In F.M.G. de~Jong and A.~Nijholt, editors, {\em Natural Language
  Interfaces, From Laboratory to Commercial and User Environments --
  Proceedings of the 5th Twente Workshop on Language Technology, Enschede,
  Twente University, NL}, June 1993.
\newblock Also MMC Preprint no.\ 13, Institute for Language Technology and
  Artificial Intelligence (ITK), Tilburg University, NL.

\bibitem{Small}
D.W. Small and L.J. Weldon.
\newblock {An Experimental Comparison of Natural and Structured Query
  Languages}.
\newblock {\em Human Factors}, 25(3):253--263, 1983.

\bibitem{SunSimplify}
Sun Microsystems Inc.
\newblock {\em {SunSimplify 2.0 Reference Manual}}, 1989.

\bibitem{Tansel3}
A.~Tansel, J.~Clifford, S.K. Gadia, S.~Jajodia, A.~Segev, and R.T. Snodgrass.
\newblock {\em {Temporal Databases -- Theory, Design, and Implementation}}.
\newblock Benjamin/Cummings, California, 1993.

\bibitem{Templeton}
M.~Templeton and J.~Burger.
\newblock Problems in {N}atural {L}anguage {I}nterface to {DBMS} with
  {E}xamples from {EUFID}.
\newblock In {\em Proceedings of the 1st Conference on Applied Natural Language
  Processing, Santa Monica, California}, pages 3--16, 1983.

\bibitem{Tennant1}
H.R. Tennant, K.M. Ross, M.~Saenz, C.W. Thompson, and J.R. Miller.
\newblock Menu-{B}ased {N}atural {L}anguage {U}nderstanding.
\newblock In {\em Proceedings of the 21st Annual Meeting of ACL, Cambridge,
  Massachusetts}, pages 151--158, 1983.

\bibitem{Tennant2}
R.~Tennant, K.M. Ross, and Thompson C.W.
\newblock Usable {N}atural {L}anguage {I}nterfaces through {M}enu-{B}ased
  {N}atural {L}anguage {U}nderstanding.
\newblock In {\em Proceedings of CHI'83, Conference on Human Factors in
  Computer Systems, Boston}. ACM, 1983.

\bibitem{Thompson2}
B.H. Thompson and F.B. Thompson.
\newblock Introducing {ASK}, {A} {S}imple {K}nowledgeable {S}ystem.
\newblock In {\em Proceedings of the 1st Conference on Applied Natural Language
  Processing, Santa Monica, California}, pages 17--24, 1983.

\bibitem{Thompson1}
B.H. Thompson and F.B. Thompson.
\newblock {ASK} is {T}ransportable in {H}alf a {D}ozen {W}ays.
\newblock {\em ACM Transactions on Office Information Systems}, 3(2):185--203,
  April 1985.

\bibitem{Thompson3}
F.B. Thompson, P.C. Lockermann, B.H. Dostert, and R.~Deverill.
\newblock {REL: A Rapidly Extensible Language System}.
\newblock In {\em {Proceedings of the 24th ACM National Converence, New York}},
  pages 399--417, 1969.

\bibitem{Thompson4}
F.B. Thompson and B.H. Thompson.
\newblock {Practical Natural Language Processing: The REL System Prototype}.
\newblock In M.~Rubinoff and M.C. Yovits, editors, {\em {Advances in
  Computers}}, pages 109--168. Academic Press, New York, 1975.

\bibitem{Intellect}
Trinzic Corporation, Bethesda, MD.
\newblock {\em {INTELLECT -- Natural Language System}}.
\newblock (Commercial leaflet).

\bibitem{KBMS}
Trinzic Corporation, Bethesda, MD.
\newblock {\em {KBMS -- Knowledge Base Management System}}.
\newblock (Commercial leaflets).

\bibitem{Ullman}
J.D. Ullman.
\newblock {\em {P}rinciples of {D}atabase and {K}nowledge-{B}ase {S}ystems--
  {V}olume 1}.
\newblock Computer Science Press, Rockville, Maryland, 1988.

\bibitem{Wallace1984}
M.~Wallace.
\newblock {\em {Communicating with Databases in Natural Language}}.
\newblock Ellis Horwood, New York, 1984.

\bibitem{Waltz}
D.L. Waltz.
\newblock {An English Language Question Answering System for a Large Relational
  Database}.
\newblock {\em Communications of the ACM}, 21(7):526--539, July 1978.

\bibitem{Warren}
D.~Warren and F.~Pereira.
\newblock An {E}fficient {E}asily {A}daptable {S}ystem for {I}nterpreting
  {N}atural {L}anguage {Q}ueries.
\newblock {\em Computational Linguistics}, 8(3-4):110--122, July-December 1982.

\bibitem{Weischedel}
R.~Weischedel.
\newblock {A Hybrid Approach to Representation in the JANUS Natural Language
  Processor}.
\newblock In {\em Proceedings of the 27th Annual Meeting of ACL, Vancouver,
  British Columbia}, pages 193--202, 1989.

\bibitem{Whittaker}
S.~Whittaker and P.~Stenton.
\newblock {User Studies and the Design of Natural Language Systems}.
\newblock In {\em Proceedings of the 4th Conference of the European Chapter of
  ACL, Manchester, England}, pages 116--123, April 1989.

\bibitem{Whittemore}
G.~Whittemore, K.~Ferrara, and H.~Brunner.
\newblock {Empirical Study of Predictive Powers of Simple Attachment Schemes
  for Post-Modifier Prepositional Phrases}.
\newblock In {\em Proceedings of the 28th Annual Meting of ACL, Pittsburgh,
  Pensylvania}, pages 23--30, June 1990.

\bibitem{Woods1968}
W.A. Woods.
\newblock {Procedural Semantics for a Question-Answering Machine}.
\newblock In {\em Proceedings of the Fall Joint Computer Conference}, pages
  457--471, New York, NY, 1968. AFIPS.

\bibitem{Woods}
W.A. Woods, R.M. Kaplan, and B.N. Webber.
\newblock The {L}unar {S}ciences {N}atural {L}anguage {I}nformation {S}ystem:
  {F}inal {R}eport.
\newblock BBN Report 2378, Bolt Beranek and Newman Inc., Cambridge,
  Massachusetts, 1972.

\end{thebibliography}

\end{document}